\documentclass[letterpaper]{article} 
\usepackage{aaai24}  
\usepackage{times}  
\usepackage{helvet}  
\usepackage{courier}  
\usepackage[hyphens]{url}  
\usepackage{graphicx} 
\usepackage{natbib}  
\usepackage{caption} 
\usepackage{algorithm}
\usepackage{algorithmic}

\usepackage{newfloat}
\usepackage{listings}
\DeclareCaptionStyle{ruled}{labelfont=normalfont,labelsep=colon,strut=off} 
\floatstyle{ruled}
\newfloat{listing}{tb}{lst}{}
\floatname{listing}{Listing}
\usepackage{makecell}
\usepackage{amsmath}
\usepackage{amssymb}

\title{Attention Guided CAM: \\Visual Explanations of Vision Transformer Guided by Self-Attention}
\author {
    Saebom Leem\textsuperscript{\rm 1,\rm 2},
    Hyunseok Seo\textsuperscript{\rm 1}\protect \thanks{Corresponding author.}
}
\affiliations {
    \textsuperscript{\rm 1}Korea Institute of Science and Technology\\
    \textsuperscript{\rm 2}Sogang University\\
    toqha1215@sogang.ac.kr, seo@kist.kr
}

\begin{document}

\maketitle

\begin{abstract}
Vision Transformer(ViT) is one of the most widely used models in the computer vision field with its great performance on various tasks. In order to fully utilize the ViT-based architecture in various applications, proper visualization methods with a decent localization performance are necessary, but these methods employed in CNN-based models are still not available in ViT due to its unique structure. In this work, we propose an attention-guided visualization method applied to ViT that provides a high-level semantic explanation for its decision. Our method selectively aggregates the gradients directly propagated from the classification output to each self-attention, collecting the contribution of image features extracted from each location of the input image. These gradients are additionally guided by the normalized self-attention scores, which are the pairwise patch correlation scores. They are used to supplement the gradients on the patch-level context information efficiently detected by the self-attention mechanism. This approach of our method provides elaborate high-level semantic explanations with great localization performance only with the class labels. As a result, our method outperforms the previous leading explainability methods of ViT in the weakly-supervised localization task and presents great capability in capturing the full instances of the target class object. Meanwhile, our method provides a visualization that faithfully explains the model, which is demonstrated in the perturbation comparison test.
\end{abstract}

\newcommand{\etal}{\textit{et al}., }
\newcommand{\ie}{\textit{i}.\textit{e}., }
\newcommand{\eg}{\textit{e}.\textit{g}.\ }
\newcolumntype{C}[1]{>{\centering\arraybackslash}m{#1}}
\newcolumntype{L}[1]{>{\arraybackslash}m{#1}}

\section{Introduction}
Transformer-based models \cite{Transformer, BERT, roberta, gpt1} is a widely used architecture in various NLP tasks due to its superior performance. Vision Transformer (ViT) \cite{ViT} is a modified Transformer that adopts the architecture of BERT \cite{BERT}, but is applicable to images by replacing its basic unit of operation with image patches. As a Transformer-based model, ViT applies the self-attention mechanism as its primary operation, sharing the advantages of the Transformer over other models: it significantly reduces the required computational load and supports better parallelization. Furthermore, recent studies propose that ViT is better at shape recognition \cite{ViTShape} and shows high robustness against occlusions and perturbations in the input \cite{ViTRobust}. 
Exploiting these benefits, ViT and its derived models have achieved remarkable performance in numerous vision tasks such as classification \cite{ CrossViT, localvit, DEIT}, object detection \cite{SwinTransformer, pyramidViT}, and semantic segmentation \cite{ViTforDense, rethinkViTSeg}. Demonstrating its high versatility and decent performance, especially in large-scale image data, it is now considered as a practical alternative to Convolutional Neural Network (CNN) \cite{ResNet, CNN, VGG, GoogLeNet} which has dominated the computer vision field for the past decade. 

Despite the notable success of ViT in computer vision, it still lacks explainability. The proper methods to provide a visual explanation of the model are vital to ensure the reliability of the given model. For CNN, for example, numerous methods have been developed to provide a faithful explanation of the model by gradient analysis \cite{hirescam, GradCAM, CAM}. In addition, many of the gradient-based methods have been actively utilized in weakly-supervised localization \cite{GradCAMpp, infoCAM, combiCAM}. In contrast, the unique structure of ViT, such as the use of $[class]$ token and the self-attention mechanism, makes it complicated to provide the proper explanation of the model. Therefore, compared to CNN, there have been fewer explainability methods developed, including Attention rollout \cite{AttentionRollout} and Layer-wise Relevance Propagation (LRP)-based method \cite{LrpForViT}.

\begin{figure}
    \normalsize
    \centering
    \begin{tabular}{p{0.2\columnwidth}p{0.20\columnwidth}p{0.22\columnwidth}p{0.17\columnwidth}}
        \makecell{Raw\\Attention}
        &
        \makecell{Attention\\Rollout}
        &
        \makecell{LRP-based}
        &
        \makecell{Ours}
    \end{tabular}
    \includegraphics[width=\columnwidth]{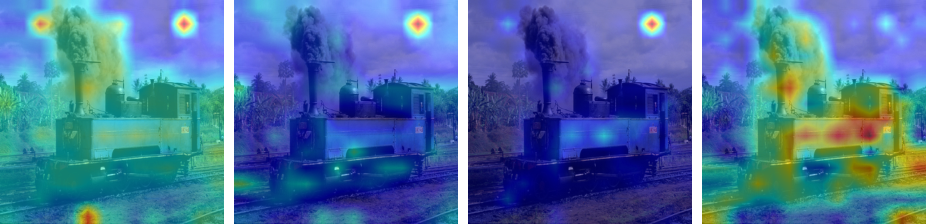}
    \includegraphics[width=\columnwidth]{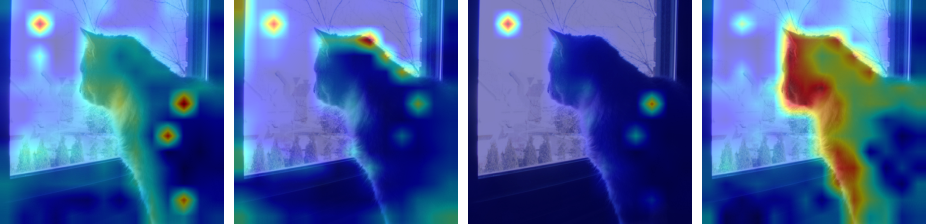}
    \caption{The illustration of peak intensity propagation from the self-attention scores to final visualization heatmaps of PASCAL VOC 2012. Raw attention is a simple sum aggregation of the self-attention scores of all layers.}
    \label{fig:peak_intensity}
\end{figure}

Attention Rollout is a method developed for ViT and aims to provide a concise aggregation of the overall attention by using the resulting matrix of self-attention operation. Although it considers the core component of ViT architecture, it assumes a linear combination of attention and overlooks the influence of the MLP head, resulting in a rough and non-class-specific explanation of the classification decision. On the other hand, the LRP-based method applied to ViT provides a class-specific analysis and takes the whole model into consideration. It focuses on decomposing the model back into the level of image patches and calculates the relevancy score of each patch based on the conservation property. 
Since both methods take the self-attention operation into account, they are prone to the peak intensity resulting from the repeated softmax operation in the sequential self-attention module. The softmax operation tends to amplify the local large values in the process of converting the self-attention scores into probabilities. Consequently, it generates a peak intensity that highlights the specific point of a homogeneous background of the input image due to high self-attention scores from similar pixel intensities. As demonstrated in Figure \ref{fig:peak_intensity}, Attention Rollout and LRP-based method are severely influenced by the peak intensity, resulting in poor localization performance. In contrast, our method renormalizes the self-attention scores with sigmoid, which does not affect the original prediction process, and therefore is much less disturbed by the peak intensity.

In this work, we propose an attention-guided gradient analysis that aims to improve localization performance by combining the essential target gradients with the feedforward feature of the self-attention module. Specifically, to provide the class activation map (CAM) of high-semantic explanation, we aggregate the gradients that are directly connected to the MLP head and backpropagated along skip connections. Also, we conclude that the self-attention score represents the patch correlation scores with a continuous pattern and preserves spatial position information. Therefore, we use the self-attention score, which is newly normalized with the sigmoid operation to alleviate peak intensities, as feature maps that guide the gradients on the pattern information of the image. In short, the proposed method provides the CAM that represents the image features of the input combined with their contributions to the prediction of the model. This approach achieves greater weakly-supervised localization performance with the state-of-the-art result in most evaluation benchmarks. The contributions of this work are as follows:

\begin{itemize}
    \item We propose a gradient-based method applicable to Vision Transformer that fully considers the major structures of the model and provides a reliable high-semantic explanation of the model.
    \item The proposed method aggregates the selective gradients guided by the self-attention to construct a class activation map (CAM) of great localization performance. 
    \item Our method outperforms the previous leading methods applied to ViT in the experiments on weakly-supervised localization. We also demonstrate the improved reliability of our method by pixel perturbation experiment. 
\end{itemize} 

\section{Related Works}

Explainability of a deep neural network matters because the black box nature of it makes it difficult to ensure that the model is working in a proper way. Hence, there have been various methods that aim to explain the model's inner workings, but each method adopts a different idea of what it intends to explain and how it generates the explanations. 
For example, in Attention Rollout \cite{AttentionRollout} which is designed to explain the Transformer, the explanation means the amount of information propagated from the first to the last self-attention module. Although it can be easily applied to any Transformer-based model, it does not take the MLP heads into account and cannot specify how much each captured correlation contributes to the classification output of a particular class. Therefore, Attention Rollout produces a non-class-specific explanation and shows lower performance in localization tasks for some regions that are unrelated to the classification output are also highlighted.

LRP-based methods are contrived to calculate the relevancy score of the input pixel to the classification output. In other words, the explanation provided by LRP is the contribution of each pixel of the input image throughout the model from the input to the output. It first decomposes a model pixel-wisely typically using Deep Taylor Decomposition (DTD) framework \cite{DTD}, then it calculates the relevancy of each of the pixels by propagating the decomposed relevancies backward from the output to the input layer. LRP-based methods have been extended to various models. Bach \etal \cite{LRPpixel} proposed an LRP method that can consider the nonlinearity of the model, and Binder \etal \cite{LRPDNN} applied LRP to some deep neural networks including GoogLeNet and VGG. They, then, extend LRP to the renormalization layer \cite{LRPnorm}. Finally, Chefer \etal \cite{LrpForViT} introduced the LRP-based method applied to ViT by proposing the method to apply LRP to the GELU \cite{GELU} layer, skip connections, and matrix multiplication, which are the major operations of ViT and calculates the relevancy score of each image patch. These methods capture the contribution of each independent and discrete unit of the model and provide a precise explanation. However, they often result in scattered contributions which only highlight a partial area of the target class object because of approximation error in relevancy calculation and incomplete attention scores as shown in Figure \ref{fig:peak_intensity}.

\begin{figure*}[!]
\begin{center}
\includegraphics[width=\textwidth]{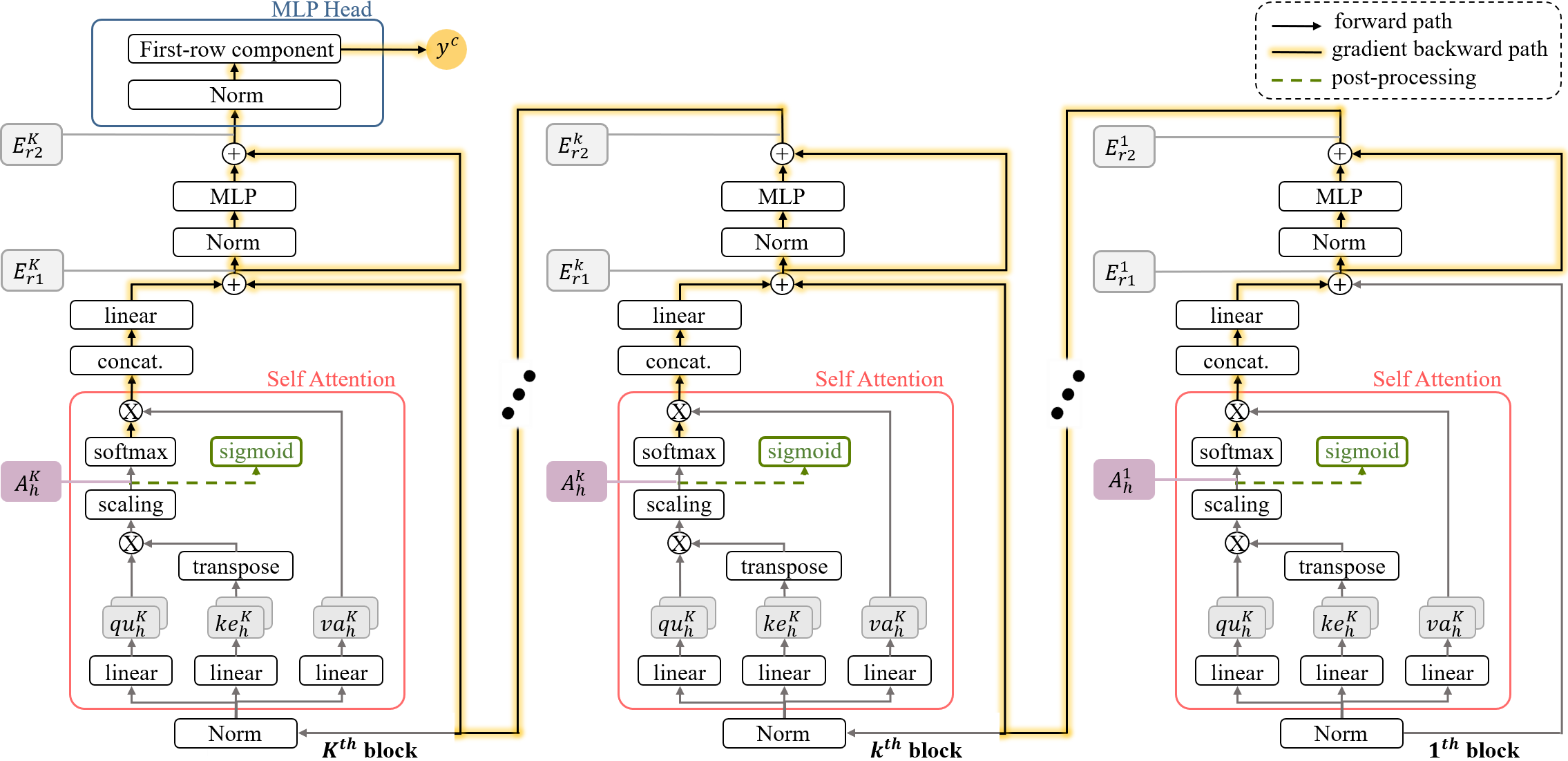}
\end{center}
   \caption{The demonstration of the ViT architecture and the major components in our method. The yellow shaded lines represent the essential gradients being considered along the skip connections propagated from the classification output of the given class $c$, $y^c$. The purple-colored boxes point to the self-attention score matrices which are the result of matrix multiplication of the query and the key matrices. The feature maps are these self-attention score matrices normalized with sigmoid, which are represented as the green boxes in each block. These feature maps are aggregated with the gradients to provide the final class activation map.}
   \label{fig:map}
\end{figure*}

On the other hand, the gradient-based methods provide a high-semantic explanation of the model, meaning that they explain the contribution of the image features elicited through multiple layers, rather than the contribution of the independent pixels. The earliest gradient-based method is Class Activation Map (CAM) \cite{CAM}, which generates a saliency map as a result of the weighted sum of the feature map channels of the last convolutional layer. Here, the weights for each feature map channel are calculated by a single time backward from the classification output of the target class. Grad-CAM \cite{GradCAM} proposes a generalized CAM, whose usage is not restricted to a model with Global Average Pooling (GAP). Grad-CAM averages the gradients of each channel to construct a class activation map of a general CNN model with fully connected layers. HiResCAM \cite{hirescam} provides a more faithful explanation by replacing the channel-wise weights of the Grad-CAM with the pixel-wise multiplication of the gradients and the feature maps. Gradient-based methods are also highly utilized in weakly-supervised object localization with great performance. Grad-CAM++ \cite{GradCAMpp} proposed a generalized version of Grad-CAM with improved localization performance, adding channel-wise weights to the Grad-CAM. Combinational CAM \cite{combiCAM} and infoCAM \cite{infoCAM} integrate the CAM of non-label classes to localize the target object more precisely. In these gradient-based methods, the feature maps reflect the interaction among the pixels from multiple layers of the model and the gradients are the contributions of these high-level image features. This approach of the gradient-based methods that combines the feature maps and their gradients intrinsically results in a continuous heatmap where contributions cluster together on the target object and also gives an excellent capability in object localization.

\section{Methodology}

Our method applies the gradient-based visualization technique to ViT \cite{ViT} to generate the class activation map (CAM) \cite{CAM, GradCAM} of the target class. The major components of our method are demonstrated in Figure \ref{fig:map}. To generate a high-semantic explanation of the model, we focus on the gradients from the classification output to each encoder block along the backward path through the skip connection. In addition, these essential gradients are guided by feature maps obtained from the newly normalized self-attention score matrices by sigmoid. The reasons why the gradients and the feature map are obtained from the self-attention blocks are as follows. Firstly, the attention score matrice at each block contains the high-level image features elicited through the self-attention mechanism. Given that these image features represent pairwise patch correlations, these matrices are appropriate to be used as a feature map. Secondly, regardless of the aggregation option chosen in the MLP head (\eg $[class]$ token or average pooling), these matrices preserve the patch position information of the input image. Here in this paper, we explain our method based on the original ViT model with a $[class]$ token.

According to the ViT architecture, the input with a size of $[(n \times p) \times (n \times p) \times 3]$ is flattened and converted into a patch embedding with a size of $[N \times P]$ before fed into the transformer encoder. Here, the number of patches $N$ equals $n^2+1$ with $n^2$ image patches and 1 additional for $[class]$ token, and the patch embedding size $P$ can vary but is generally defined as $[p^2 \times 3]$. The order of the patches embedded in this process is maintained and therefore the positional information of the $n^2$ image patches is traceable throughout the feedforward encoder block.

ViT architecture can be largely divided into the MLP head and the encoder blocks which consist of the self-attention, MLP, and skip connections. The patch embedding shape of $[N \times P]$ is maintained at every skip connection layer and we denote this matrix after the first skip connection in the $k^{th}$ encoder block as $E_{r1}^{k}$ and the one after the second skip connection in the $k^{th}$ encoder block as $E_{r2}^{k}$, respectively. During the multi-head self-attention operation in the $k^{th}$ encoder, the matrix multiplication of the query and the key matrices results in the self-attention score matrices with the size of $[H \times N \times N]$ where $H$ is the number of heads, and we denote the self-attention matrix of $h^{th}$ head as $A_{h}^{k} (1 \leq h \leq H)$. At the end of the encoder, the MLP head produces the classification outputs and we denote the classification output of the target class $c$ as $y^c$. These matrices in the ViT feedforward network and their notations are demonstrated in Figure \ref{fig:map}.

\begin{figure}
    \normalsize
    \centering
    \begin{tabular}{C{0.02\columnwidth}C{0.9\columnwidth}}
        \rotatebox{90}{softmax} &
        \includegraphics[width=0.9\columnwidth]{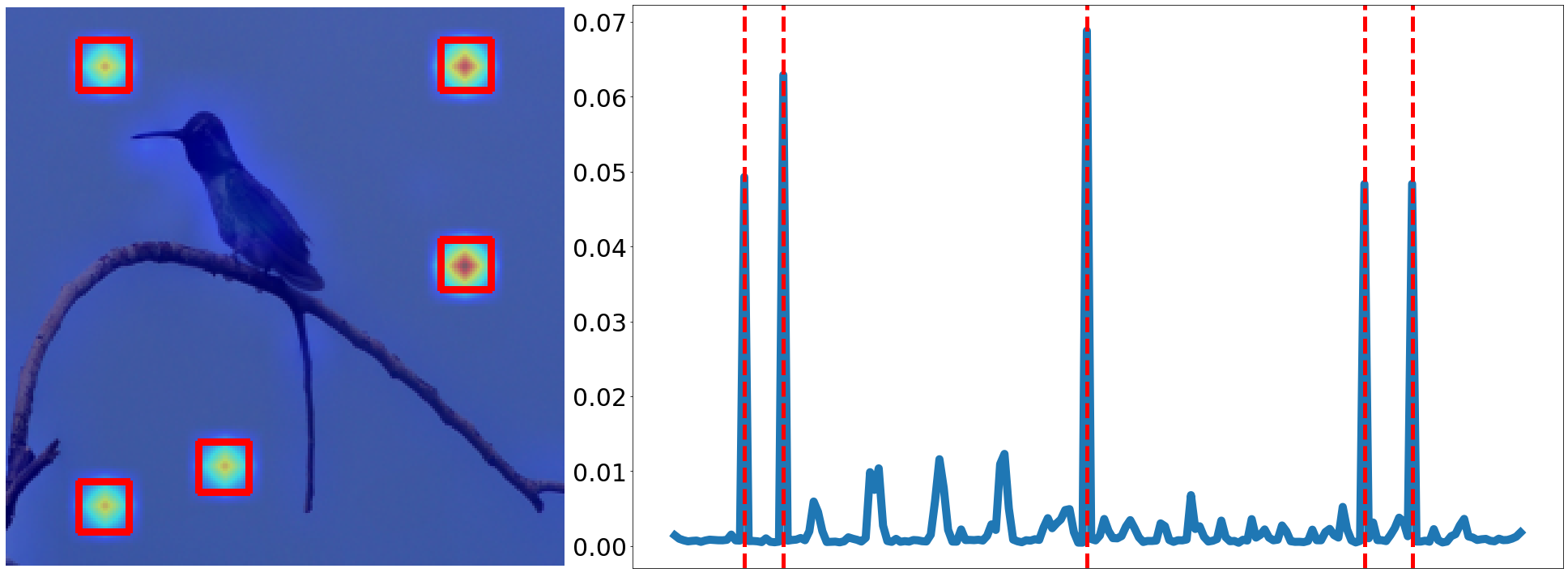}
        \\
        \rotatebox{90}{sigmoid}
        &
        \includegraphics[width=0.9\columnwidth]{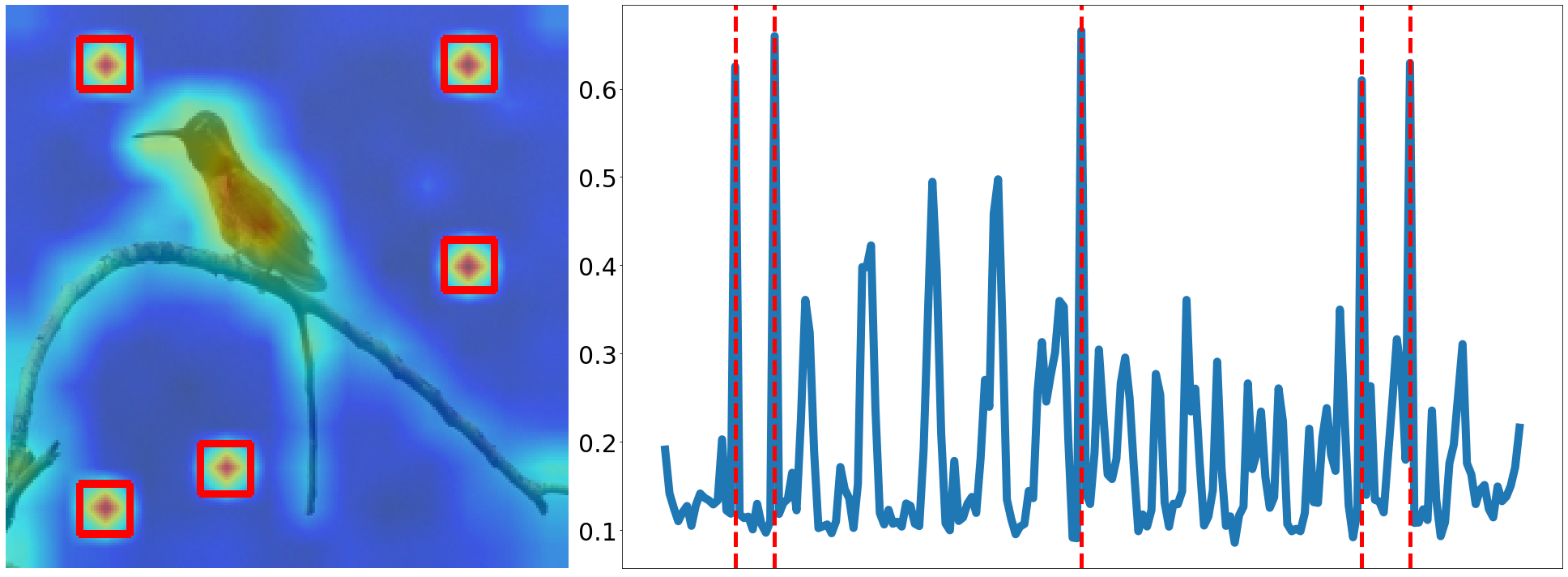}
    \end{tabular}
    \caption{The demonstration of the results of softmax and sigmoid operation applied on self-attention. The images (left) are the sum aggregation of self-attention of all layers and heads with each operation and the peaks are indicated with red boxes.
    The graphs (right) are the distributions of the flattened self-attention scores in the left images. The peaks are indicated with red lines.}
    \label{fig:sigmoid}
\end{figure}

As the feature map for gradient calculation, the self-attention matrices $A_{h}^{k}$ are used. Each element of the matrices represents the pairwise patch correlation scores detected at each layer and head and can guide the combined gradients on meaningful pattern information. Basically, in ViT, the self-attention scores are converted into the probability by the softmax operation. However, softmax tends to maximize the local large values and generates some peak intensity that suppresses other important values as shown in Figure \ref{fig:sigmoid}. Therefore, instead of softmax, we normalize the self-attention matrices with sigmoid, which is a monotonically increasing function as well as softmax. When we denote the softmax function as $S(\cdot)$ and sigmoid function as $G(\cdot)$, the two function satisfies the following relation:
\begin{equation}
    \forall x, y \in \mathbb{R}, S(x)<S(y) \implies G(x)<G(y)
\end{equation}
At the same time, the sigmoid effectively recovers the medium correlations that are lost in softmax. The effect of replacing softmax with sigmoid is represented in Figure \ref{fig:sigmoid}. To prevent misunderstanding, we clarify that normalization with sigmoid does not affect any backpropagation process of the original ViT structure, and sigmoid of $A_{h}^{k}$ is calculated after the model finishes learning.

Note that in ViT with $[class]$ token, only the first rows of $E_{r1}^{k}$, $E_{r2}^{k}$ and $A_{h}^{k}$ (\ie $E_{r1,1}^{k}$, $E_{r2,1}^{k}$ and $A_{h,1}^{k}$, respectively) are considered. The MLP head is only connected with the $[class]$ token at the end of the last encoder block, $E_{r2,1}^{K}$, which does not contain the positional information of the image patches itself. However, the positional information connected to this token can be traced back to the first rows of the self-attention matrices $A_{h,1}^{k}$s since all operations from $A_{h, i}^{k}$s along the skip connections are applied row-wisely where $i$ stands for $i^{th}$ row component. Also due to the skip connection, the MLP head is directly connected not only to $A_{h,1}^{K}$ in the last encoder block but also to all $A_{h,1}^{k}$s in the previous blocks, as shown in Figure \ref{fig:map}. Therefore, the feature map $F_{h}^{k}$ in our method consists of the first-row components of $A_{h,i}^{k}$s normalized with sigmoid operation and is defined as:
\begin{equation}
    F_{h}^k = G(A_{h,1}^k)
\end{equation}
The feature maps $F_{h}^k$s are generated in the green boxes in Figure \ref{fig:map}.

To produce a complete CAM, the gradients, which represent the influence of $A_{h,1}^{k}$s of each block, should be combined with the feature maps. In the $k^{th}$ encoder block except for the last encoder block (\ie, $k < K$), the gradient directly connected from the MLP head towards $E_{r1,1}^{k}$ is propagated along the first skip connection in the $(k+1)^{th}$ block towards $E_{r2,1}^{k}$. By doing so, the gradient from the MLP head can be propagated to $E_{h,1}^{k-1}$ as well as to $E_{h,1}^{k}$. The skip connection consists of a residual operation, a simple addition of two matrices, with no effect on the gradient passing through it. Therefore, for the first skip connection in the $(k + 1)^{th}$ encoder block connected to $E_{r2,1}^{k}$, we get the mathematical relation as follows:
\begin{equation}
    \frac{\partial E_{r1,1}^{k+1}}{\partial E_{r2,1}^{k}} = I
    \label{eqn: one}
\end{equation}
Then let us denote the gradient propagated from the output $y^c$ to the matrix $E_{r1,1}^{k}$ in the $k^{th}$ encoder block along the skip connection path as $\beta^{k,c}$. The gradient $\beta^{k,c}$ is defined as:
\begin{equation}
    \beta ^{k, c} = \begin{cases} \frac{\partial y^c}{\partial E_{r1,1}^k}, &k=K
    \\\beta^{k+1,c}\cdot\frac{\partial E_{r1,1}^{k+1}}{\partial E_{r2,1}^k}\cdot\frac{\partial E_{r2,1}^k}{\partial E_{r1,1}^k}, &k<K
    \end{cases}
    \label{eqn: two}
\end{equation}
From Eqs. \ref{eqn: one} and \ref{eqn: two}, we can get:
\begin{equation}
    \beta ^{k, c} = \begin{cases} \frac{\partial y^c}{\partial E_{r1,1}^k}, &k=K
    \\\beta^{k+1,c}\cdot\frac{\partial E_{r2,1}^k}{\partial E_{r1,1}^k}, &k<K
    \end{cases}
    \label{eqn: three}
\end{equation}

Since the self-attention matrices pass through a softmax layer in each block, the gradients $\alpha_h^{k,c}$ that are propagated to each feature map $F_h^k$ are defined as:
\begin{equation}
    \alpha_{h}^{k,c} = \beta^{k,c}\cdot\frac{\partial E_{r1,1}^k}{\partial F_h^k} = \beta^{k,c}\cdot\frac{\partial E_{r1,1}^k}{\partial S(A_{h,1}^k)}\cdot\frac{\partial S(A_{h,1}^k)}{\partial G(A_{h,1}^k)}
    \label{eqn: alpha prime}
\end{equation}
However, the gradients still cause the peak-amplification effect since they contain the weights propagated from softmax. In other words, the peak amplification effect occurs due to large elements in  $\frac{\partial S(A_{h,1}^k)}{\partial G(A_{h,1}^k)}$.
However, if the general attention scores are assumed to have a smooth varying property (\ie $\frac{\partial S(A_{h,1}^k)}{\partial G(A_{h,1}^k)} \thickapprox 1$), the gradients propagated to each $S(A_{h,1}^k)$ approximate the gradients propagated to each $G(A_{h,1}^k)$, which can be formulated as:
\begin{equation}
    \alpha_h^{k,c} = \beta^{k,c}\cdot\frac{\partial E_{r1,1}^k}{\partial S(A_{h,1}^k)}\cdot I
    \label{eqn: four}
\end{equation}
and it is proved in the supplementary material.
Due to Equation \ref{eqn: four}, our method still faithfully explains the model. The final gradients $\alpha_h^{k,c}$ are demonstrated in the yellow-shaded line in Figure \ref{fig:map}.

\begin{figure} [!]
    \centering
    \includegraphics[width=0.8\columnwidth]{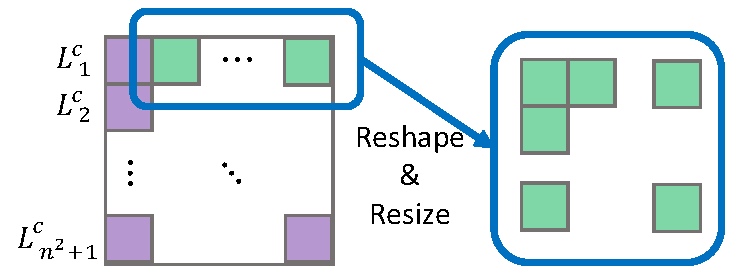}
    \caption{The illustration of how the one-dimensional matrix $L^c_1$ is reshaped into a two-dimensional class activation map.}
    \label{fig:position}
\end{figure}

Finally, the class activation map $L^c$ of the given class $c$ can be formulated as:
\begin{equation}
    L^c = \sum_{k=1}^{K}\sum_{h=1}^{H}F_{h}^k\odot ReLU(\alpha_{h}^{k,c})
\end{equation}
where $\odot$ refers to the Hadamard product. Here, we apply the ReLU operation on the computed gradient $\alpha_{h}^{k,c}$ to reflect only the positive contribution to the classification output. Also, the contributions obtained at each location of the patch from all layers and heads are summed to combine them in the same way as the feedforward network fuses the embedded patches at each skip connection. The result of this process, $L^c$, is a $[1\times N]$ matrix where each value represents the contribution of each patch to the classification output of the given class $c$. However, the first element of this matrix represents the contribution of the $[class]$ token which does not contain any spatial information. Since we are only interested in the contribution of each image patch, we discard the first element and construct the class activation map (CAM) with the last $n^2$ elements. To visualize the final CAM, the $n^2$ elements are reshaped to a two-dimensional image, which has the size of $[n \times n]$ as demonstrated in Figure. \ref{fig:position}. Then it is interpolated to have the same size as the input image and eventually generates the final CAM of the model.

\section{Experiments}

In this section, we present the results of the performance comparison of our method with previous leading methods. The compared methods here are the current explainability methods devised for ViT that consist of Attention Rollout \cite{AttentionRollout} and LRP-based method for ViT \cite{LrpForViT}. 

\subsection{Experimental Setup}
\textbf{Datasets and Evaluation Metrics.} For the evaluation, we used the validation set of ImageNet ILSVRC 2012 \cite{ILSVRC} and Pascal VOC 2012 \cite{pascal} and the test set of Caltech-UCSD Birds-200-2011 (CUB 200) \cite{CUB200}, which provide the bounding-box annotation label. In quantitative evaluation, the images with more than one class label in PASCAL VOC 2012 are excluded and only single-class images are used. During the weakly-supervised localization evaluation, the input images for which the model produces a wrong prediction are excluded since the heatmaps are not reliable in this case.

For the weakly-supervised localization test, the performance is measured by pixel accuracy, Intersection over Union (IoU), Dice coefficient (F1), precision, and recall scores. The pixel perturbation test is measured by the ABPC score \cite{AOPC} with pixel-level perturbation. The ABPC score is the area between the LeRF and MoRF perturbation curves where the LeRF curve removes the least relevant pixels first and the MoRF curve removes the most relevant pixels first. A larger ABPC value indicates a better quality of the heatmap.

\noindent\textbf{Implementation Details.} All methods are evaluated with the same ViT-base \cite{ViT} model that takes the input image with a size of $[224 \times 224 \times 3]$. All methods share the same model parameters and the fine-tuning details of the model parameters are provided in the supplementary material. In this ViT, the input images are converted into $[14 \times 14]$ number of patches and therefore each method generates a heatmap with a size of $[14 \times 14 \times 1]$ where one pixel corresponds to the contribution of one image patch of the input image. Before evaluation, the heatmaps are all resized into $[224 \times 224 \times 1]$ and adjusted to a min-max normalization. For the weakly-supervised object detection, we get a binary mask from the generated heatmap by applying a threshold ($\sigma=0.5$) and then generate bounding boxes from the group of pixels that have a continuous contour. The perturbation test is applied to the ground-truth class to compare the heatmap quality on the existing object.

\subsection{Results}
Here we demonstrate the visualization of the heatmaps generated by all three methods. Then, we present the quantitative evaluation of each method measured by the weakly-supervised object detection metrics and pixel-perturbation.

\begin{figure*}[!th]
  \begin{center}
  \normalsize
  \begin{tabular}{L{0.3in}C{2.in}|C{2.in}|C{2.in}}
    {} & \makecell{ImageNet ILSVRC 2012} & \makecell{Pascal VOC 2012} & \makecell{CUB 200}
    \\\rotatebox{90}{\makecell{Bounding\\Box}}
    &\includegraphics[width=\linewidth]{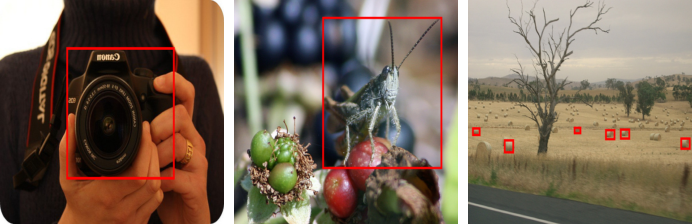}
    &\includegraphics[width=\linewidth]{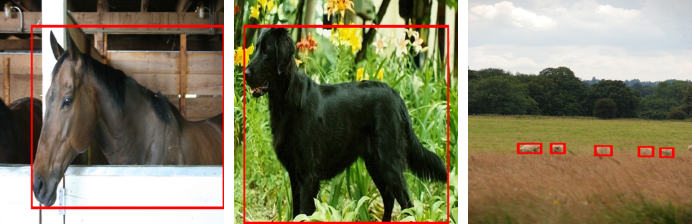}
    &\includegraphics[width=\linewidth]{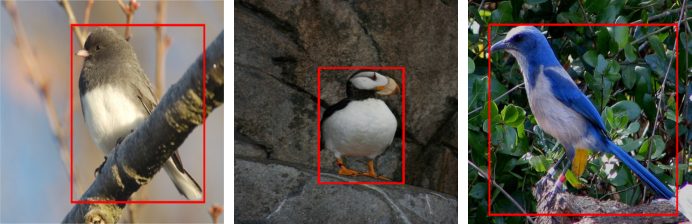}\\
    \rotatebox{90}{\makecell{Attention\\Rollout}}
    &\includegraphics[width=\linewidth]{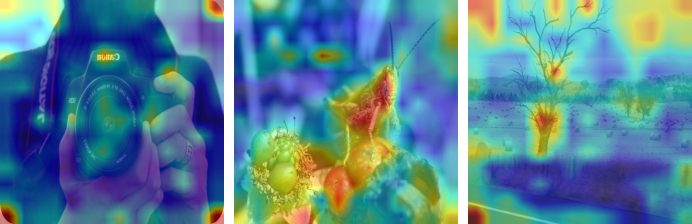}
    &\includegraphics[width=\linewidth]{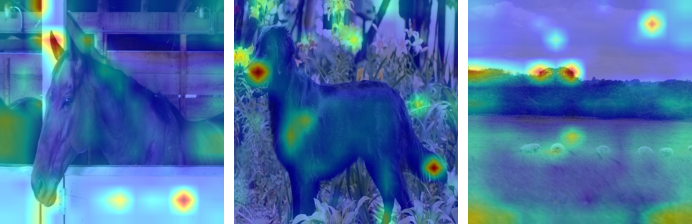}
    & \includegraphics[width=\linewidth]{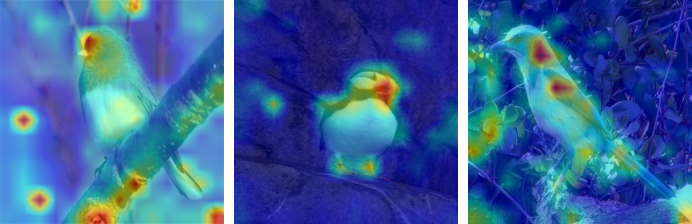}\\
    \rotatebox{90}{\makecell{LRP-based}}
    &\includegraphics[width=\linewidth]{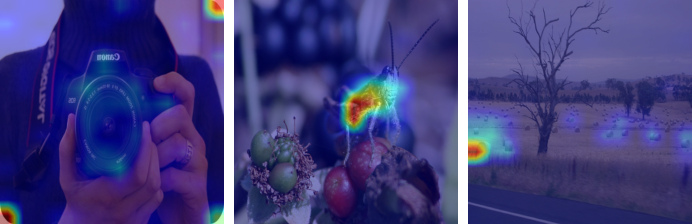} 
    &\includegraphics[width=\linewidth]{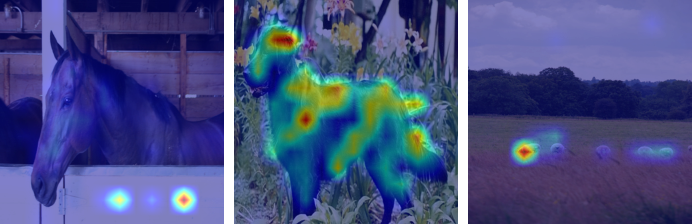}
    &\includegraphics[width=\linewidth]{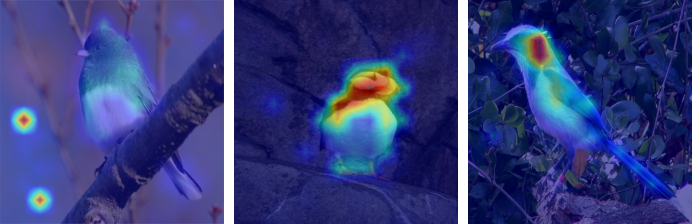}\\
    \rotatebox{90}{\makecell{Ours}}
    &\includegraphics[width=\linewidth]{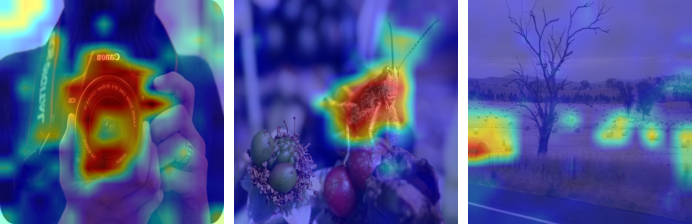} 
    &\includegraphics[width=\linewidth]{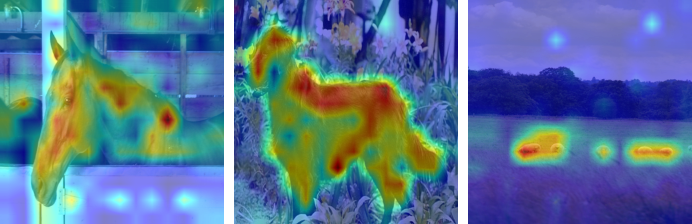}
    &\includegraphics[width=\linewidth]{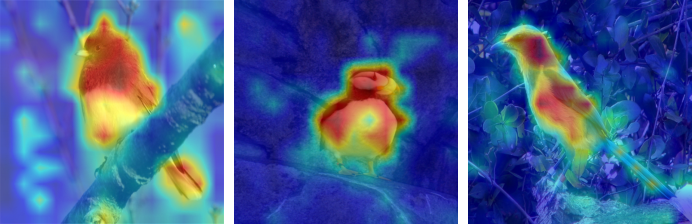}
  \end{tabular}
  \end{center}
  \caption{The heatmaps on ImageNet ILSVRC 2012, Pascal VOC 2012, and CUB 200 dataset generated by each of the methods. The first images in each dataset demonstrate the peak intensities generated on a homogeneous non-object background in Attention Rollout and LRP-based method and the reduced peak intensities in our method. The second and third images in ILSVRC 2012 and PASCAL VOC show the localization performance of each method on single-instance and multiple-instance images, respectively. CUB200 consists of single-instance images only and its second and third images include one object instance per image.}
  \label{fig:heatmap_localization}
\end{figure*}

\begin{figure}[!t]
  \begin{center}
  \normalsize
  \begin{tabular}{p{0.36in}p{0.6in}p{0.63in}p{0.60in}}
       {} & \makecell{Attention\\Rollout} & \makecell{LRP-based} & \makecell{Ours} \\
  \end{tabular}
  \begin{tabular}{p{0.4in}C{2.39in}}
      Cat & \includegraphics[width=\linewidth]{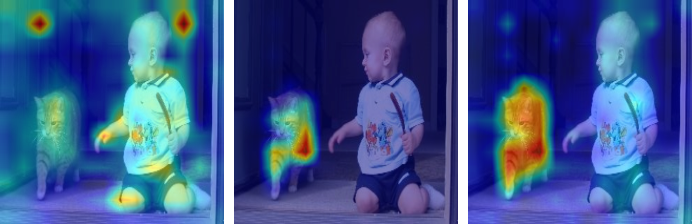} \\
      Person& \includegraphics[width=\linewidth]{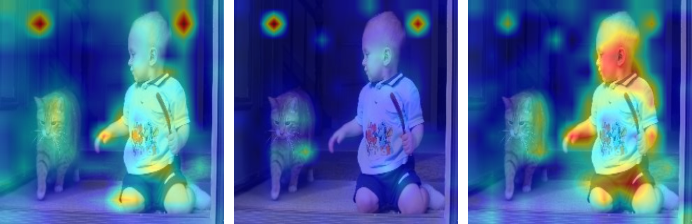} \\
      Car & \includegraphics[width=\linewidth]{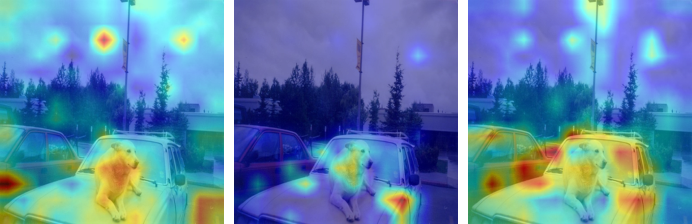} \\
      Dog & \includegraphics[width=\linewidth]{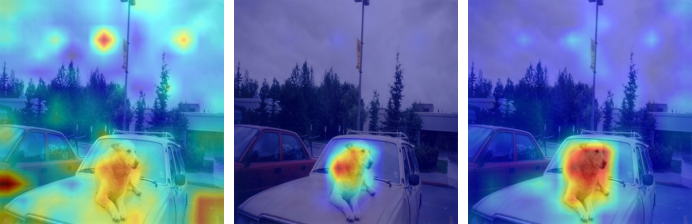}
  \end{tabular}
  \end{center}
    \caption{Visualization of the heatmaps generated for different target objects. The input images are from PASCAL VOC 2012 and have two class labels per image.}
    \label{fig:class specific}
\end{figure}

\noindent\textbf{Visualization.}
The visualization results of each method on three datasets are presented in Figure \ref{fig:heatmap_localization}. The first images of each dataset demonstrate that our method greatly improves object localization performance by successfully shading the peak intensities. As can be seen, the visualization results of the Attention Rollout and LRP-based method on the images are dominated by the peak intensities and therefore fail to localize the target objects which are indicated by the bounding boxes. Also, this result shows that our method captures the object region more precisely compared to Attention Rollout \cite{AttentionRollout}. Since Attention Rollout produces a non-class-specific explanation, it often highlights some unrelated background regions whereas our method successfully separates the foreground object from the background. 
Also, our method shows a strong point in encompassing the full object area compared to the LRP-based method \cite{LrpForViT}. The LRP-based method tends to highlight a small conspicuous part of the object. Furthermore, it often misses some instances of the object in an image with multiple instances of the class object. In contrast, our method successfully localizes the whole object and also captures all the instances of the class object even when the instances are placed far from each other. 

In addition, ours can also provide class-specific explanations for different classes presented within an image. Figure \ref{fig:class specific} demonstrates the results of the explanation with different target classes generated by each method. Attention Rollout does not provide a class-specific explanation and therefore produces the same results regardless of the target class label.  In contrast, LRP and ours are both capable of providing a class-specific explanation of the given model. Here, our method still maintains better localization performance by fully capturing the object of each target class. The visualization results on more image samples are presented in the supplementary material.

\noindent\textbf{Weakly-Supervised Object Detection.}  
The result of the weakly-supervised object detection on the ImageNet ILSVRC 2012 validation set is presented in Table \ref{Tab:ILSVRC}. This result shows that ours achieves 73.41\% in pixel accuracy, 52.12\% in IoU score, and 65.15\% in dice coefficient, which is the highest among the three methods. Although there was a drop in precision compared to LRP-based (82.99\% vs 91.10\%), ours achieves a much better recall score (62.76\% vs 21.76\%) which is generally a trade-off with precision.

\begin{table}[!t]
\small
\centering

\begin{tabular}{l|ccc}
\hline 
               & \makecell{Attention Rollout} & LRP-based  & Ours            \\ \hline \hline
pixel accuracy & 0.6209           & 0.5863          & \textbf{0.7341} \\
IoU            & 0.3597            & 0.2029          & \textbf{0.5212} \\
dice (F1)      & 0.4893           & 0.3055         & \textbf{0.6515} \\
precision      & 0.7326           & \textbf{0.9110} & 0.8299          \\
recall         & 0.4657           & 0.2176          & \textbf{0.6276} \\ \hline
\end{tabular}%
\caption{Localization performance comparison on ImageNet ILSVRC 2012.}
\label{Tab:ILSVRC}
\end{table}

\begin{table}[!t]
\small
\centering

\begin{tabular}{l|ccc}
\hline
               & \makecell{Attention Rollout}  & LRP-based     & Ours           \\ \hline \hline
pixel accuracy & 0.5592          & 0.5750          & \textbf{0.7521} \\
IoU            & 0.1645          & 0.1574          & \textbf{0.5335} \\
dice (F1)      & 0.2431          & 0.2348          & \textbf{0.6646} \\
precision      & 0.5802          & 0.7651          & \textbf{0.8167} \\
recall         & 0.2115          & 0.1716          & \textbf{0.6647} \\ \hline
\end{tabular}%
\caption{Localization performance comparison on Pascal VOC 2012.}
\label{Tab:Pascal}

\end{table}

\begin{table}[!t]
\small
\centering
\begin{tabular}{l|ccc}
\hline
               & \makecell{Attention Rollout}  & LRP-based             & Ours            \\ \hline \hline
pixel accuracy & 0.7273            & 0.7039          & \textbf{0.8351} \\
IoU            & 0.3097            & 0.1997          & \textbf{0.5836} \\
dice (F1)      & 0.4339            & 0.3106          & \textbf{0.7220} \\
precision      & 0.8357            & \textbf{0.9669} & 0.8987          \\
recall         & 0.3420            & 0.1992          & \textbf{0.6438} \\ \hline
\end{tabular}%
\caption{Localization performance comparison on CUB 200.}
\label{Tab:CUB}

\end{table}

\begin{table}[!t]
\small
\centering
\begin{tabular}{l|ccc}
\hline
               & Attention Rollout  & LRP-based             & Ours            \\ \hline \hline
LeRF               & 0.4739            & 0.5140          & \textbf{0.5298} \\
MoRF               & 0.2053            & 0.1736          & \textbf{0.1607} \\
ABPC               & 0.2685            & 0.3404          & \textbf{0.3691} \\
\hline

\end{tabular}%
\caption{The result of pixel perturbation test on ILSVRC 2012. LeRF represents the area under the LeRF curve and MoRF represents the area under the MoRF curve. The ABPC score is the area between the LeRF and MoRF curves. For LeRF and ABPC higher is better and for MoRF lower is better.}
\label{Tab:perturbation}

\end{table}

The localization performance on the Pascal VOC 2012 validation is presented in Table \ref{Tab:Pascal}. Our method achieves 75.21\% in pixel accuracy, 53.35\% in IoU, and 66.46\% in dice coefficient, presenting outstanding localization performance. In this case, our method also achieves the highest precision score of 81.67\%.

Table \ref{Tab:CUB} shows the localization performance evaluated on the test set of CUB 200. The CUB 200 dataset consists of images with a single bird per image which are easily distinguishable from the background. Therefore, patch-correlation information of the self-attention scores serves a significant role in this dataset. Our method effectively utilizes the self-attention scores, resulting in a more accurate and precise explanation compared to others, achieving 83.51\% in pixel accuracy, 58.36\% in IoU, and 72.20\% in dice coefficient. This is about 10.78\%, 27.39\%, and 28.81\% higher scores respectively than those of Attention Rollout.

In conclusion, our method provides a high-semantic explanation of ViT that consistently shows an outstanding performance in the weakly-supervised object detection task compared to the Attention rollout and LRP-based method. Our method shows significant improvements in terms of pixel accuracy, IoU, recall, and dice coefficient, while it still maintains an acceptable level of precision. 

\noindent\textbf{Pixel Perturbation.}
The result of the pixel perturbation test is presented in Table \ref{Tab:perturbation}. LeRF and MoRF represent the areas under the prediction probability score curve when removing the least relevant pixels first and the most relevant pixels first, respectively. The ABPC is the area between these two curves, which is obtained by subtracting the AUC of the MoRF curve from that of the LeRF curve. Our method achieves a better LeRF score and MoRF score and therefore a higher ABPC score compared to the LRP-based method (36.91 \% vs 34.04 \%). This guarantees the better faithfulness and reliability of the explanations that our method provides. Additional evaluation results of the object localization task and the pixel perturbation test are presented in the supplementary materials.

\section{Conclusion}
In this work, we propose an attention-guided gradient analysis method that aims at achieving greater weakly-supervised localization performance. To this end, our method provides a high-level semantic explanation by selectively collecting the essential gradients propagated from the classification output of the target class to each self-attention matrix along the skip connection path. To supplement the gradient information with the patch correlation information that indicates the group of patches with contiguous patterns, the self-attention scores are combined with the gradients as feature maps. Before these two major components are aggregated, the self-attention scores are adapted in a way that decreases the effect of peak intensities to improve the localization performance of the CAM.
As a result, our method outperforms the current state-of-the-art visualization techniques of ViT by localizing the full areas of the target object, and it especially achieves a great performance improvement in capturing the multiple instances of the given class object. This provides a reliable explanation of the model and weakly-supervised object detection method at the same time and allows ViT to be more adaptable to many tasks involving object localization in the computer vision field. 

\section{Acknowledgments}
This work was supported by KIST Institutional Programs (2V09831, 2E32341, and 2E32211).

\bibliography{Leem}

\begin{thebibliography}{37}
\providecommand{\natexlab}[1]{#1}

\bibitem[{Abnar and Zuidema(2020)}]{AttentionRollout}
Abnar, S.; and Zuidema, W. 2020.
\newblock Quantifying attention flow in transformers.
\newblock \emph{arXiv preprint arXiv:2005.00928}.

\bibitem[{Bach et~al.(2015)Bach, Binder, Montavon, Klauschen, M{\"u}ller, and Samek}]{LRPpixel}
Bach, S.; Binder, A.; Montavon, G.; Klauschen, F.; M{\"u}ller, K.-R.; and Samek, W. 2015.
\newblock On pixel-wise explanations for non-linear classifier decisions by layer-wise relevance propagation.
\newblock \emph{PloS one}, 10(7): e0130140.

\bibitem[{Binder et~al.(2016{\natexlab{a}})Binder, Bach, Montavon, M{\"u}ller, and Samek}]{LRPDNN}
Binder, A.; Bach, S.; Montavon, G.; M{\"u}ller, K.-R.; and Samek, W. 2016{\natexlab{a}}.
\newblock Layer-wise relevance propagation for deep neural network architectures.
\newblock In \emph{Information science and applications (ICISA) 2016}, 913--922. Springer.

\bibitem[{Binder et~al.(2016{\natexlab{b}})Binder, Montavon, Lapuschkin, M{\"u}ller, and Samek}]{LRPnorm}
Binder, A.; Montavon, G.; Lapuschkin, S.; M{\"u}ller, K.-R.; and Samek, W. 2016{\natexlab{b}}.
\newblock Layer-wise relevance propagation for neural networks with local renormalization layers.
\newblock In \emph{Artificial Neural Networks and Machine Learning--ICANN 2016: 25th International Conference on Artificial Neural Networks, Barcelona, Spain, September 6-9, 2016, Proceedings, Part II 25}, 63--71. Springer.

\bibitem[{Chattopadhay et~al.(2018)Chattopadhay, Sarkar, Howlader, and Balasubramanian}]{GradCAMpp}
Chattopadhay, A.; Sarkar, A.; Howlader, P.; and Balasubramanian, V.~N. 2018.
\newblock Grad-cam++: Generalized gradient-based visual explanations for deep convolutional networks.
\newblock In \emph{2018 IEEE winter conference on applications of computer vision (WACV)}, 839--847. IEEE.

\bibitem[{Chefer, Gur, and Wolf(2021)}]{LrpForViT}
Chefer, H.; Gur, S.; and Wolf, L. 2021.
\newblock Transformer interpretability beyond attention visualization.
\newblock In \emph{Proceedings of the IEEE/CVF Conference on Computer Vision and Pattern Recognition}, 782--791.

\bibitem[{Chen, Fan, and Panda(2021)}]{CrossViT}
Chen, C.-F.~R.; Fan, Q.; and Panda, R. 2021.
\newblock Crossvit: Cross-attention multi-scale vision transformer for image classification.
\newblock In \emph{Proceedings of the IEEE/CVF international conference on computer vision}, 357--366.

\bibitem[{Devlin et~al.(2018)Devlin, Chang, Lee, and Toutanova}]{BERT}
Devlin, J.; Chang, M.-W.; Lee, K.; and Toutanova, K. 2018.
\newblock Bert: Pre-training of deep bidirectional transformers for language understanding.
\newblock \emph{arXiv preprint arXiv:1810.04805}.

\bibitem[{Dosovitskiy et~al.(2020)Dosovitskiy, Beyer, Kolesnikov, Weissenborn, Zhai, Unterthiner, Dehghani, Minderer, Heigold, Gelly et~al.}]{ViT}
Dosovitskiy, A.; Beyer, L.; Kolesnikov, A.; Weissenborn, D.; Zhai, X.; Unterthiner, T.; Dehghani, M.; Minderer, M.; Heigold, G.; Gelly, S.; et~al. 2020.
\newblock An image is worth 16x16 words: Transformers for image recognition at scale.
\newblock \emph{arXiv preprint arXiv:2010.11929}.

\bibitem[{Draelos and Carin(2020)}]{hirescam}
Draelos, R.~L.; and Carin, L. 2020.
\newblock Hirescam: Faithful location representation in visual attention for explainable 3d medical image classification.
\newblock \emph{arXiv preprint arXiv:2011.08891}.

\bibitem[{Everingham et~al.(2012)Everingham, Van~Gool, Williams, Winn, and Zisserman}]{pascal}
Everingham, M.; Van~Gool, L.; Williams, C. K.~I.; Winn, J.; and Zisserman, A. 2012.
\newblock The {PASCAL} {V}isual {O}bject {C}lasses {C}hallenge 2012 {(VOC2012)} {R}esults.
\newblock http://www.pascal-network.org/challenges/VOC/voc2012/workshop/index.html.

\bibitem[{He et~al.(2016)He, Zhang, Ren, and Sun}]{ResNet}
He, K.; Zhang, X.; Ren, S.; and Sun, J. 2016.
\newblock Deep residual learning for image recognition.
\newblock In \emph{Proceedings of the IEEE conference on computer vision and pattern recognition}, 770--778.

\bibitem[{Hendrycks and Gimpel(2016)}]{GELU}
Hendrycks, D.; and Gimpel, K. 2016.
\newblock Gaussian error linear units (gelus).
\newblock \emph{arXiv preprint arXiv:1606.08415}.

\bibitem[{LeCun et~al.(1989)LeCun, Boser, Denker, Henderson, Howard, Hubbard, and Jackel}]{CNN}
LeCun, Y.; Boser, B.; Denker, J.~S.; Henderson, D.; Howard, R.~E.; Hubbard, W.; and Jackel, L.~D. 1989.
\newblock Backpropagation applied to handwritten zip code recognition.
\newblock \emph{Neural computation}, 1(4): 541--551.

\bibitem[{Li et~al.(2021)Li, Zhang, Cao, Timofte, and Van~Gool}]{localvit}
Li, Y.; Zhang, K.; Cao, J.; Timofte, R.; and Van~Gool, L. 2021.
\newblock Localvit: Bringing locality to vision transformers.
\newblock \emph{arXiv preprint arXiv:2104.05707}.

\bibitem[{Liu et~al.(2019)Liu, Ott, Goyal, Du, Joshi, Chen, Levy, Lewis, Zettlemoyer, and Stoyanov}]{roberta}
Liu, Y.; Ott, M.; Goyal, N.; Du, J.; Joshi, M.; Chen, D.; Levy, O.; Lewis, M.; Zettlemoyer, L.; and Stoyanov, V. 2019.
\newblock Roberta: A robustly optimized bert pretraining approach.
\newblock \emph{arXiv preprint arXiv:1907.11692}.

\bibitem[{Liu et~al.(2021)Liu, Lin, Cao, Hu, Wei, Zhang, Lin, and Guo}]{SwinTransformer}
Liu, Z.; Lin, Y.; Cao, Y.; Hu, H.; Wei, Y.; Zhang, Z.; Lin, S.; and Guo, B. 2021.
\newblock Swin transformer: Hierarchical vision transformer using shifted windows.
\newblock In \emph{Proceedings of the IEEE/CVF international conference on computer vision}, 10012--10022.

\bibitem[{Montavon et~al.(2017)Montavon, Lapuschkin, Binder, Samek, and M{\"u}ller}]{DTD}
Montavon, G.; Lapuschkin, S.; Binder, A.; Samek, W.; and M{\"u}ller, K.-R. 2017.
\newblock Explaining nonlinear classification decisions with deep taylor decomposition.
\newblock \emph{Pattern recognition}, 65: 211--222.

\bibitem[{Naseer et~al.(2021)Naseer, Ranasinghe, Khan, Hayat, Shahbaz~Khan, and Yang}]{ViTRobust}
Naseer, M.~M.; Ranasinghe, K.; Khan, S.~H.; Hayat, M.; Shahbaz~Khan, F.; and Yang, M.-H. 2021.
\newblock Intriguing properties of vision transformers.
\newblock \emph{Advances in Neural Information Processing Systems}, 34: 23296--23308.

\bibitem[{Qin, Kim, and Gedeon(2021)}]{infoCAM}
Qin, Z.; Kim, D.; and Gedeon, T. 2021.
\newblock Informative Class Activation Maps.
\newblock \emph{arXiv preprint arXiv:2106.10472}.

\bibitem[{Radford et~al.(2018)Radford, Narasimhan, Salimans, Sutskever et~al.}]{gpt1}
Radford, A.; Narasimhan, K.; Salimans, T.; Sutskever, I.; et~al. 2018.
\newblock Improving language understanding by generative pre-training.

\bibitem[{Ranftl, Bochkovskiy, and Koltun(2021)}]{ViTforDense}
Ranftl, R.; Bochkovskiy, A.; and Koltun, V. 2021.
\newblock Vision transformers for dense prediction.
\newblock In \emph{Proceedings of the IEEE/CVF International Conference on Computer Vision}, 12179--12188.

\bibitem[{Russakovsky et~al.(2015)Russakovsky, Deng, Su, Krause, Satheesh, Ma, Huang, Karpathy, Khosla, Bernstein, Berg, and Fei-Fei}]{ILSVRC}
Russakovsky, O.; Deng, J.; Su, H.; Krause, J.; Satheesh, S.; Ma, S.; Huang, Z.; Karpathy, A.; Khosla, A.; Bernstein, M.; Berg, A.~C.; and Fei-Fei, L. 2015.
\newblock {ImageNet Large Scale Visual Recognition Challenge}.
\newblock \emph{International Journal of Computer Vision (IJCV)}, 115(3): 211--252.

\bibitem[{Samek et~al.(2016)Samek, Binder, Montavon, Lapuschkin, and M{\"u}ller}]{AOPC}
Samek, W.; Binder, A.; Montavon, G.; Lapuschkin, S.; and M{\"u}ller, K.-R. 2016.
\newblock Evaluating the visualization of what a deep neural network has learned.
\newblock \emph{IEEE transactions on neural networks and learning systems}, 28(11): 2660--2673.

\bibitem[{Selvaraju et~al.(2017)Selvaraju, Cogswell, Das, Vedantam, Parikh, and Batra}]{GradCAM}
Selvaraju, R.~R.; Cogswell, M.; Das, A.; Vedantam, R.; Parikh, D.; and Batra, D. 2017.
\newblock Grad-CAM: Visual Explanations From Deep Networks via Gradient-Based Localization.
\newblock In \emph{Proceedings of the IEEE International Conference on Computer Vision (ICCV)}.

\bibitem[{Simonyan and Zisserman(2014)}]{VGG}
Simonyan, K.; and Zisserman, A. 2014.
\newblock Very deep convolutional networks for large-scale image recognition.
\newblock \emph{arXiv preprint arXiv:1409.1556}.

\bibitem[{Szegedy et~al.(2015)Szegedy, Liu, Jia, Sermanet, Reed, Anguelov, Erhan, Vanhoucke, and Rabinovich}]{GoogLeNet}
Szegedy, C.; Liu, W.; Jia, Y.; Sermanet, P.; Reed, S.; Anguelov, D.; Erhan, D.; Vanhoucke, V.; and Rabinovich, A. 2015.
\newblock Going deeper with convolutions.
\newblock In \emph{Proceedings of the IEEE conference on computer vision and pattern recognition}, 1--9.

\bibitem[{Touvron et~al.(2021)Touvron, Cord, Douze, Massa, Sablayrolles, and J{\'e}gou}]{DEIT}
Touvron, H.; Cord, M.; Douze, M.; Massa, F.; Sablayrolles, A.; and J{\'e}gou, H. 2021.
\newblock Training data-efficient image transformers \& distillation through attention.
\newblock In \emph{International conference on machine learning}, 10347--10357. PMLR.

\bibitem[{Tuli et~al.(2021)Tuli, Dasgupta, Grant, and Griffiths}]{ViTShape}
Tuli, S.; Dasgupta, I.; Grant, E.; and Griffiths, T.~L. 2021.
\newblock Are convolutional neural networks or transformers more like human vision?
\newblock \emph{arXiv preprint arXiv:2105.07197}.

\bibitem[{Vaswani et~al.(2017)Vaswani, Shazeer, Parmar, Uszkoreit, Jones, Gomez, Kaiser, and Polosukhin}]{Transformer}
Vaswani, A.; Shazeer, N.; Parmar, N.; Uszkoreit, J.; Jones, L.; Gomez, A.~N.; Kaiser, {\L}.; and Polosukhin, I. 2017.
\newblock Attention is all you need.
\newblock \emph{Advances in neural information processing systems}, 30.

\bibitem[{Voita et~al.(2019)Voita, Talbot, Moiseev, Sennrich, and Titov}]{partial-lrp}
Voita, E.; Talbot, D.; Moiseev, F.; Sennrich, R.; and Titov, I. 2019.
\newblock Analyzing multi-head self-attention: Specialized heads do the heavy lifting, the rest can be pruned.
\newblock \emph{arXiv preprint arXiv:1905.09418}.

\bibitem[{Wah et~al.(2011)Wah, Branson, Welinder, Perona, and Belongie}]{CUB200}
Wah, C.; Branson, S.; Welinder, P.; Perona, P.; and Belongie, S. 2011.
\newblock The {C}altech-{UCSD} {B}irds-200-2011 {D}ataset.
\newblock Technical Report CNS-TR-2011-001, California Institute of Technology.

\bibitem[{Wang et~al.(2021)Wang, Xie, Li, Fan, Song, Liang, Lu, Luo, and Shao}]{pyramidViT}
Wang, W.; Xie, E.; Li, X.; Fan, D.-P.; Song, K.; Liang, D.; Lu, T.; Luo, P.; and Shao, L. 2021.
\newblock Pyramid Vision Transformer: A Versatile Backbone for Dense Prediction Without Convolutions.
\newblock In \emph{Proceedings of the IEEE/CVF International Conference on Computer Vision (ICCV)}, 568--578.

\bibitem[{Wightman(2019)}]{timm}
Wightman, R. 2019.
\newblock PyTorch Image Models.
\newblock \url{https://github.com/rwightman/pytorch-image-models}.

\bibitem[{Yang et~al.(2020)Yang, Kim, Kim, and Kim}]{combiCAM}
Yang, S.; Kim, Y.; Kim, Y.; and Kim, C. 2020.
\newblock Combinational class activation maps for weakly supervised object localization.
\newblock In \emph{Proceedings of the IEEE/CVF Winter Conference on Applications of Computer Vision}, 2941--2949.

\bibitem[{Zheng et~al.(2021)Zheng, Lu, Zhao, Zhu, Luo, Wang, Fu, Feng, Xiang, Torr et~al.}]{rethinkViTSeg}
Zheng, S.; Lu, J.; Zhao, H.; Zhu, X.; Luo, Z.; Wang, Y.; Fu, Y.; Feng, J.; Xiang, T.; Torr, P.~H.; et~al. 2021.
\newblock Rethinking semantic segmentation from a sequence-to-sequence perspective with transformers.
\newblock In \emph{Proceedings of the IEEE/CVF conference on computer vision and pattern recognition}, 6881--6890.

\bibitem[{Zhou et~al.(2016)Zhou, Khosla, Lapedriza, Oliva, and Torralba}]{CAM}
Zhou, B.; Khosla, A.; Lapedriza, A.; Oliva, A.; and Torralba, A. 2016.
\newblock Learning deep features for discriminative localization.
\newblock In \emph{Proceedings of the IEEE conference on computer vision and pattern recognition}, 2921--2929.

\end{thebibliography}

\newpage ~ \newpage
\twocolumn[\centering \huge \textbf{\linebreak ~ \linebreak Supplementary Materials
\linebreak ~ \linebreak ~ \linebreak ~ \linebreak ~ \linebreak}]

\section{Supplementary Proof}
Here, we prove that the gradient considered in our method, $\alpha_h^{k,c}$, is the approximation of the gradient propagated to the feature map, $\alpha_h^{'k,c}$. Let us denote the softmax function as $S(\cdot)$ and the sigmoid function as $G(\cdot)$. Then, the gradient to the original feature map, $\alpha_h^{'k,c}$, is defined as:
\begin{equation}
    \alpha_{h}^{'k,c} = \beta^{k,c}\frac{\partial E_{r1,1}^k}{\partial S(A_{h,1}^k)}\frac{\partial S(A_{h,1}^k)}{\partial G(A_{h,1}^k)}
    \label{eqn: alpha prime_}
\end{equation}
when we denote the gradients to the matrices at the first skip connection at each $k^{th}$ encoder block, $E_{r1,1}^{k}$, as  $\beta^{k,c}$ and the first-row self-attention score component in this block as $A_{h,1}^k$. As stated in the methodology section, the gradients $\alpha_{h}^{'k,c}$ include the peak-amplification from softmax due to the last term, $\frac{\partial S(A_{h,1}^k)}{\partial G(A_{h,1}^k)}$. Therefore, we redefine the gradient in our method in a way that avoids the peak-amplification effect as follows:
\begin{equation}
    \alpha_h^{k,c} = \beta^{k,c}\frac{\partial E_{r1,1}^k}{\partial S(A_{h,1}^k)}
    \label{eqn: gradient}
\end{equation}

Despite the adaptation in the gradients, the difference between the two gradients, $\alpha_h^{k,c}$ and $\alpha_h^{'k,c}$ is trivial. According to the chain rule, the gradient $\alpha_h^{'k,c}$ can be rewritten as follows:
\begin{align}
    \nonumber\alpha_{h,1}^{'k,c}&=\alpha_h^{k,c}\cdot\frac{\partial S(A_{h,1}^k)}{\partial G(A_{h,1}^k)}\\
    &=\alpha_h^{k,c}\cdot\frac{\partial S(A_{h,1}^k)}{\partial A_{h,1}^k}\cdot \left ({\frac{\partial G(A_{h,1}^k)}{\partial A_{h,1}^k}}\right )^{-1}
    \label{eqn: 6}
\end{align}
Let us denote the $j^{th}$ element that constitutes $A_{h,1}^k$ as $v_{j}$, and the softmax function on this vector as $S_{j}$. Then we can write the softmax function on the vector $v_{j}$ as follows:
\begin{equation}
    S_j = \frac{e^{v_{j}}}{\sum_{m=1}^{N}e^{v_{m}}}
\end{equation}
The derivative of $S_j$ can be written as follows: 
\begin{equation}
    \frac{\partial {S_j}}{\partial v_{j}} = 
\begin{cases}
S_j\left ( 1-S_j \right ) &\text{if } j=m
\\-S_j\cdot S_m &\text{if } j\neq m
\end{cases} 
\end{equation}
When the sigmoid operation is also applied on the same sized vector $v_j$ and we denote the sigmoid function on this vector as $G_j$, the derivative of the  $G_j$ can be written as follows:
\begin{equation}
    \frac{\partial G_j}{\partial v_j} = 
\begin{cases}
G_j\left (1-G_j \right ) &\text{if } j=m
\\ 0 &\text{if } j \neq m
\end{cases} 
\end{equation}
Then, the latter term of Equation \ref{eqn: 6} can be written as follows:
\begin{gather}
   \nonumber\frac{\partial S(A_h^k)}{\partial A_h^k}\cdot\left ({\frac{\partial G(A_h^k)}{\partial A_h^k}}\right )^{-1}
    =\\\begin{bmatrix}
\frac{S_1(1-S_1)}{G_1(1-G_1)} & -\frac{S_1S_2}{G_2(1-G_2)} & \cdots & -\frac{S_1S_{N}}{G_{N}(1-G_{N})}\\
-\frac{S_2S_1}{G_1(1-G_1)} & \frac{S_2(1-S_2)}{G_2(1-G_2)} & \cdots & -\frac{S_2S_{N}}{G_{N}(1-G_{N})} \\
\vdots & \vdots & \ddots & \vdots \\
-\frac{S_{N}S_1}{G_1(1-G_1)} & -\frac{S_{N}S_2}{G_2(1-G_2)} & \cdots & \frac{S_{N}(1-S_{N})}{G_{N}(1-G_{N})} \\
\end{bmatrix}
\label{eqn:10}
\end{gather}
This matrix can be approximated to a diagonal matrix since
\begin{equation}
    S_j(1-S_j) >> S_jS_{j'}
\end{equation}
Let us denote the $j^{th}$ element of $\alpha_h^{k,c}$ as $\alpha_{h,j}^{k,c}$ and the $j_{th}$ element of $\alpha_{h}^{'k,c}$ as $\alpha_{h,j}^{'k,c}$. Then, from Equation \ref{eqn: 6} and the diagonal matrix in Equation \ref{eqn:10}, $\alpha_{h,j}^{'k,c}$ can be rewritten as follows:

\begin{align}
    \alpha_{h, j}^{'k,c} = \alpha_{h, j}^{k,c} \cdot \frac{S_j(1-S_j)}{G_j(1-G_j)}
    \label{eqn:twelve}
\end{align}

Then, we can specify the latter term  $\frac{S_j(1-S_j)}{G_j(1-G_j)}$ in Equation \ref{eqn:twelve} as the factor that amplifies the peak intensities. When we assume that the self-attention scores on general images have the smooth varying property like power law \cite{power}, we can assume the following equation:
\begin{equation}
    \frac{S_j(1-S_j)}{G_j(1-G_j)} \thickapprox I
    \label{eqn:smooth varying}
\end{equation}
Due to Equation \ref{eqn:smooth varying}, the follow relationship is satisfied:
\begin{equation}
    \alpha_h^{k,c} \thickapprox \alpha_h^{'k,c}
    \label{eqn: approx}
\end{equation}
and therefore the gradients of our method still faithfully explain the contribution of each patch in the feature map $F_h^k$.

\begin{table*}[!t]
\normalsize
\centering
\begin{tabular}{l|ccccc}
\hline
               & \makecell{raw attention} & Grad-CAM  & full-LRP  & partial-LRP & Ours         \\ \hline \hline
pixel accuracy & 0.5079                     & 0.5138    & 0.4994    & 0.5959      & \textbf{0.7341} \\
IoU            & 0.0885                     & 0.0652    & 0.2557    & 0.2303      & \textbf{0.5212} \\
dice (F1)      & 0.1422                     & 0.1080    & 0.3215    & 0.3388      & \textbf{0.6515} \\
precision      & 0.4765                     & 0.6035    & 0.4890    & \textbf{0.8303}      & 0.8299 \\
recall         & 0.1078                     & 0.0713    & 0.5115    & 0.2543      & \textbf{0.6276} \\
ABPC           & 0.1901                     & 0.0693    & 0.0149    & 0.2700      & \textbf{0.3691} \\
\hline
\end{tabular}%
\caption{Performance comparison with additional methods on ImageNet ILSVRC 2012. The results shows the localization performance and ABPC score of each method.}
\label{Tab:other}

\end{table*}

\begin{table}[!t]
\normalsize
\begin{center}
\begin{tabular}{l|ccc}
\hline
               & \makecell{Attention\\Rollout}  & LRP-based             & Ours            \\ \hline \hline
pixel accuracy & 0.6946            & 0.7323          & \textbf{0.7477} \\
IoU            & 0.0982            & 0.1215          & \textbf{0.3746} \\
dice (F1)      & 0.1551            & 0.1863          & \textbf{0.5005} \\
precision      & 0.3058            & 0.5503          & \textbf{0.5823} \\
recall         & 0.1712            & 0.1469          & \textbf{0.6448} \\ \hline
\end{tabular}%
\end{center}
\caption{Localization performance comparison on multi-class images of Pascal VOC 2012.}
\label{Tab:multi}

\end{table}

\section{Training Details}
For the ImageNet ILSVRC 2012 \cite{ILSVRC}, we loaded the pretrained parameters provided by the timm library \cite{timm} without additional training. For the other two datasets, Pascal VOC 2012 \cite{pascal} and CUB200 \cite{CUB200}, the models are trained with SGD applying weight decay of 0.0001 and momentum of 0.9. The learning rate starts at 0.0003 and decreases by a factor of 0.95 at each epoch. Specifically, for Pascal VOC 2012, we trained the model for 60 epochs with batch size of 16. For CUB200, due to the low classification performance in shorter training sessions in this fine-grained dataset, the model was trained for 500 epochs with a batch size of 64. The quantitative performance evaluation and visualization comparison were conducted using the ViT-base model with the common parameters across all compared Explainable AI (XAI) methods.

\section{More Results}
Here, we present the results of additional evaluations in the weakly-supervised object localization task and pixel perturbation test. The implementation details and evaluation settings are identical to those used previously. Firstly, Table \ref{Tab:other} presents the comparison results with additional methods. The methods compared include raw attention, Grad-CAM \cite{GradCAM}, full-LRP \cite{LRPnorm} and partial-LRP \cite{partial-lrp}. The raw attention represents a simple average aggregation of the raw attention score matrices from all layers and heads. The result highlights the superior performance of our method in terms of pixel accuracy, IoU, dice (F1) recall and ABPC scores. Despite a slight drop in the precision, our method maintains an acceptable level of precision, exhibiting only a small gap compared to partial-LRP.

Our method, along with the LRP-based method \cite{LrpForViT}, is capable of generating class-specific explanation for the model's decisions, in contrast to Attention Rollout \cite{AttentionRollout}. Therefore, it is valuable to compare our method and the LRP-based method in multi-class images to asses their ability to distinctly localize the target image rather than capturing the most prominent random object in the given image. In this context, a multi-class image refers to an image that contains two or more class labels and corresponding objects within an image, and the heatmaps are generated from the classification output of each class except Attention Rollout. Since Attention Rollout can only produce one visualization per image, the results of Attention Rollout were redundantly evaluated, with one heatmap being assessed multiple times based on the number of classes presented in each image. 

The corresponding results on multi-class images in Pascal VOC 2012 \cite{pascal} are presented in the Table \ref{Tab:multi}. For this evaluation, 1986 multi-class images in the validation set of Pascal VOC 2012 with 4514 labels were assessed. The results demonstrate the constantly better localization performance of our method in images which contain multiple objects with different classes.

\section{More Visualizations}
Here we demonstrate more visual comparisons of each of the methods. The results of sample images from ImageNet ILSVRC 2012 are presented in Figure \ref{fig:ILSVRC_vis}, those of sample images from Pascal VOC 2012 are presented in Figure \ref{fig:Pascal_vis}, and those of sample images from CUB200 are presented in Figure \ref{fig:CUB200_vis}. The red bounding box in the bounding box image indicates the target object and the heatmpas of each method represents the visualization generated without any thresholding. 

The visualization result consistently shows that our method outperforms others in localizing the complete and multiple instances of the target object and successfully mitigates the peak intensities that often dominate the heatmaps in other methods.

\newcolumntype{C}[1]{>{\centering\arraybackslash}m{#1}}
\newcolumntype{L}[1]{>{\arraybackslash}m{#1}}
\begin{figure*}[!t]
  \begin{center}
  \small
  \begin{tabular}{C{0.65in}p{0.65in}C{0.65in}C{0.65in}L{0.03in}p{0.65in}p{0.65in}C{0.65in}C{0.65in}}
        \makecell{bounding\\box} & \makecell{Attention\\Rollout} &  LRP-based & Ours & {} & \makecell{bounding\\box} & \makecell{Attention\\Rollout} & LRP-based & Ours
  \end{tabular}
  \begin{tabular}{C{3.3in}C{3.3in}}
      \includegraphics[width=0.98\linewidth]{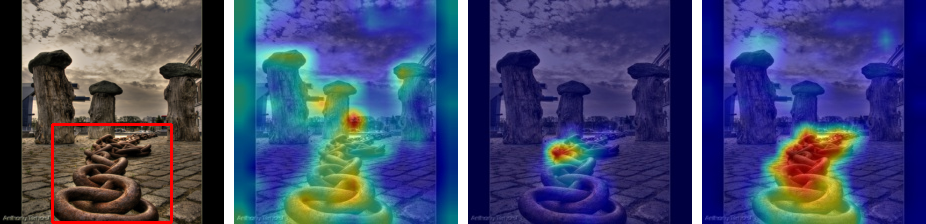} &
      \includegraphics[width=0.98\linewidth]{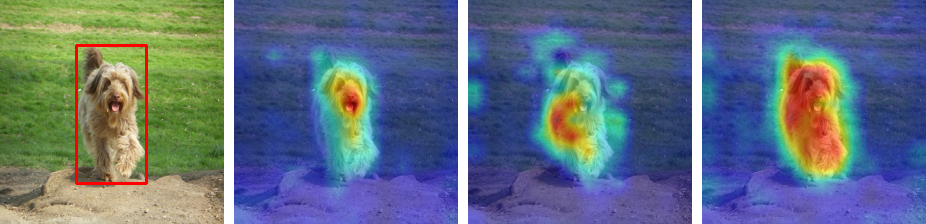}\\
      \includegraphics[width=0.98\linewidth]{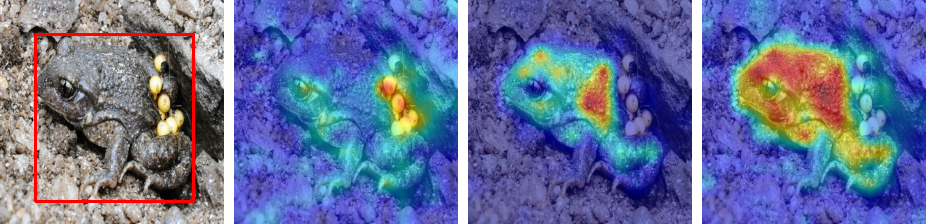} &
      \includegraphics[width=0.98\linewidth]{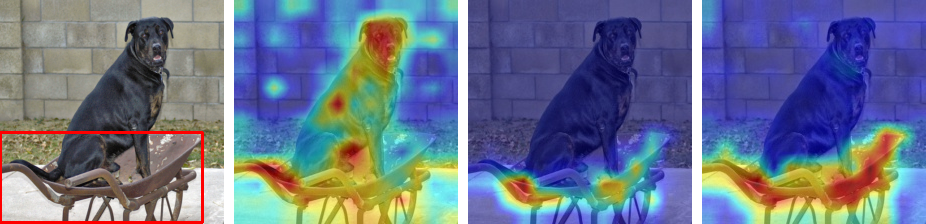}\\
      \includegraphics[width=0.98\linewidth]{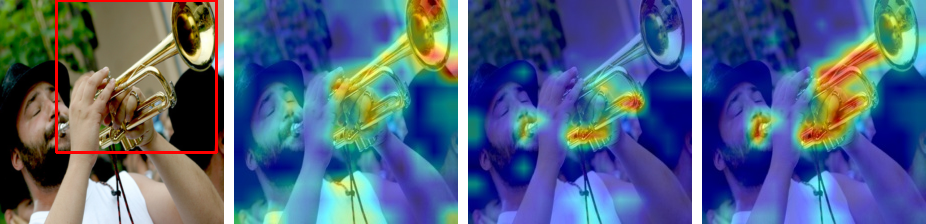} &
      \includegraphics[width=0.98\linewidth]{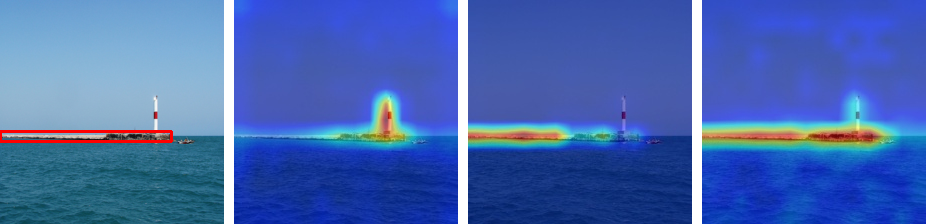}\\
      \includegraphics[width=0.98\linewidth]{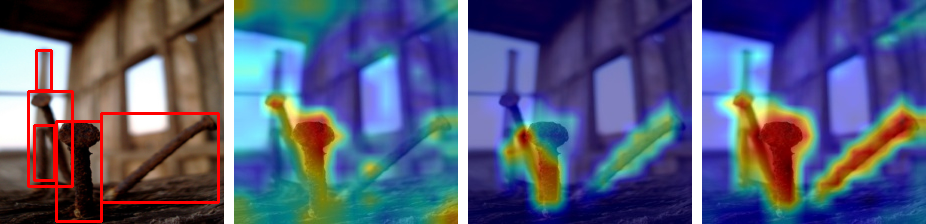} &
      \includegraphics[width=0.98\linewidth]{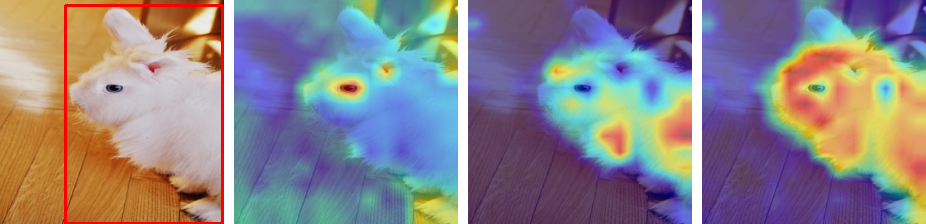}\\
      \includegraphics[width=0.98\linewidth]{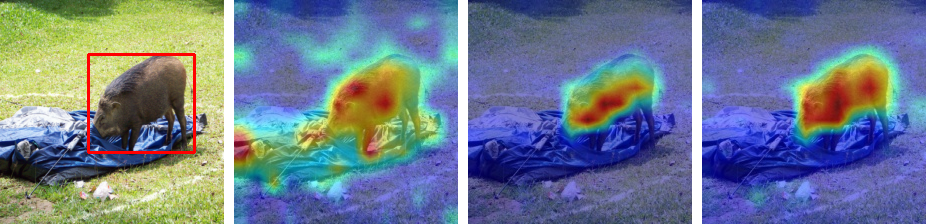} &
      \includegraphics[width=0.98\linewidth]{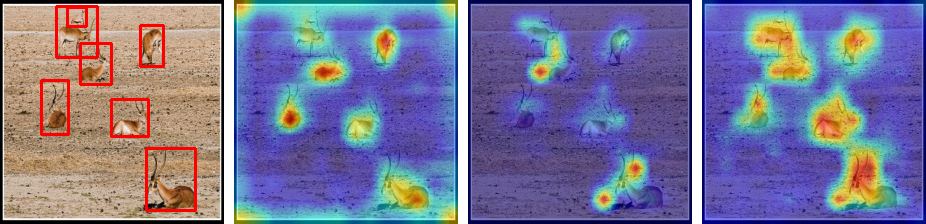}\\
      \includegraphics[width=0.98\linewidth]{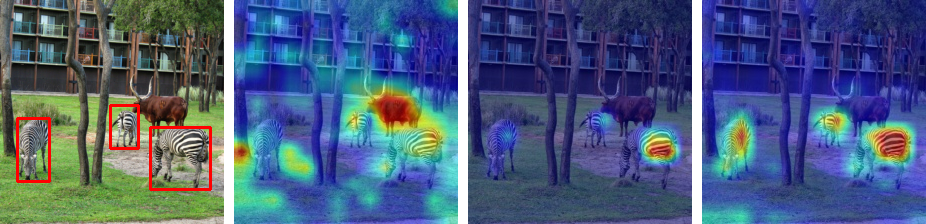} &
      \includegraphics[width=0.98\linewidth]{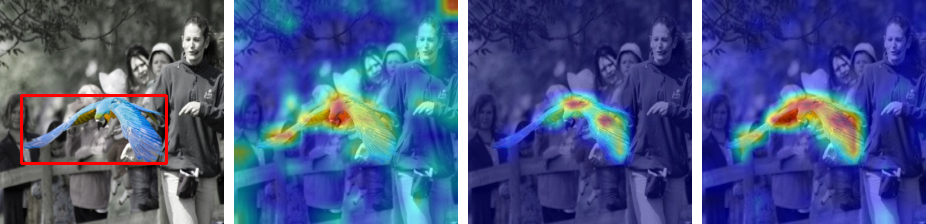}\\
      \includegraphics[width=0.98\linewidth]{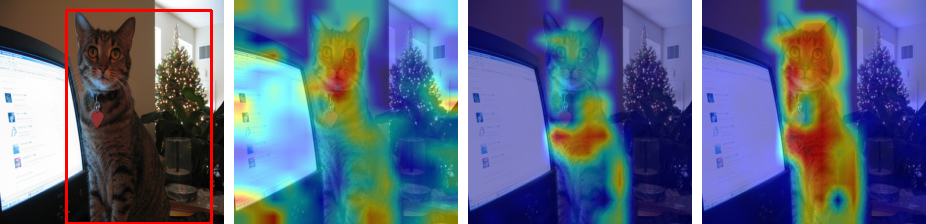} &
      \includegraphics[width=0.98\linewidth]{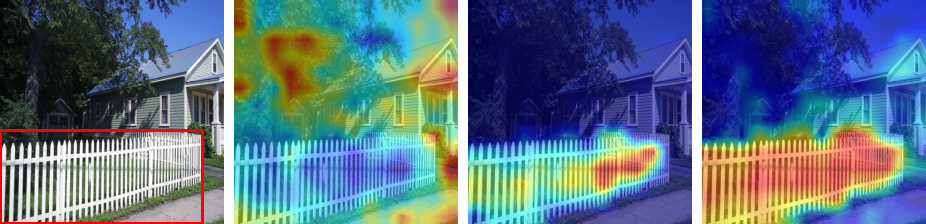}\\
      \includegraphics[width=0.98\linewidth]{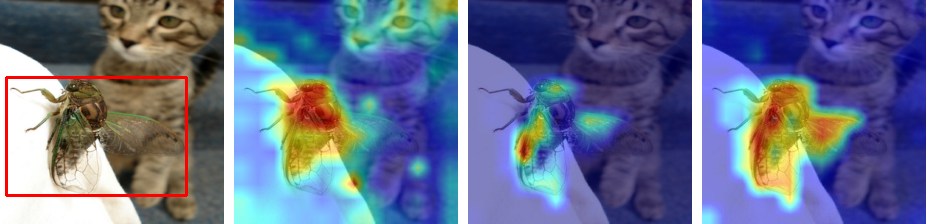} &
      \includegraphics[width=0.98\linewidth]{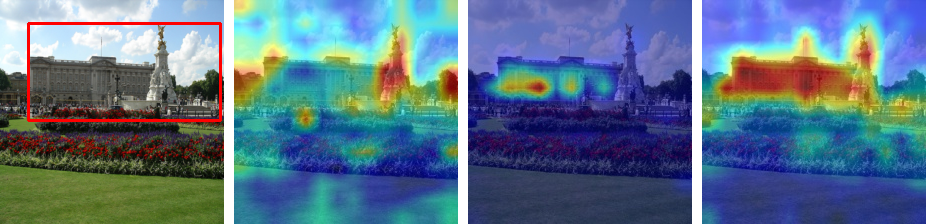}\\
      \includegraphics[width=0.98\linewidth]{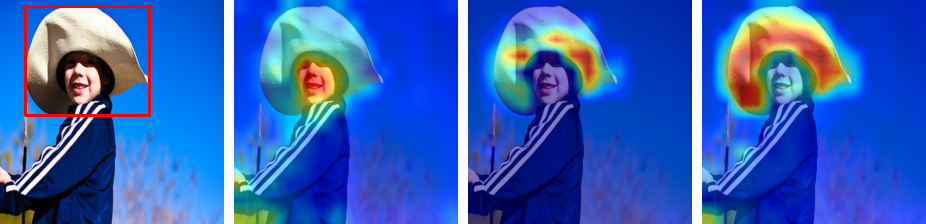} &
      \includegraphics[width=0.98\linewidth]{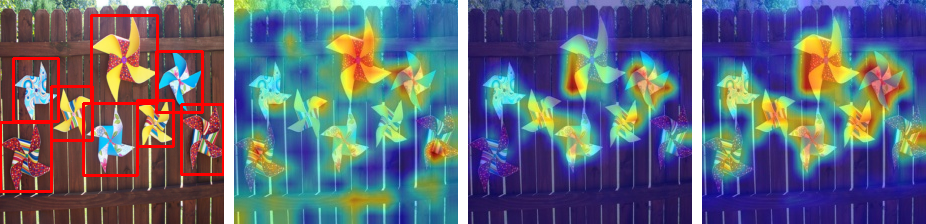}\\
      \includegraphics[width=0.98\linewidth]{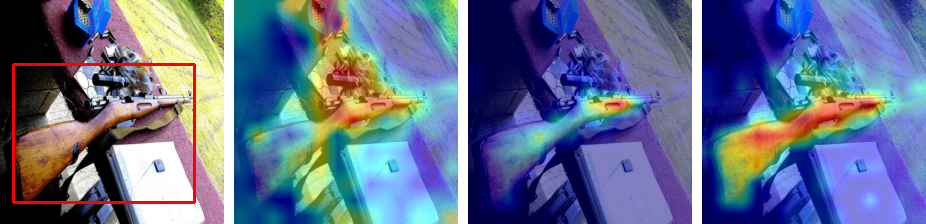} &
      \includegraphics[width=0.98\linewidth]{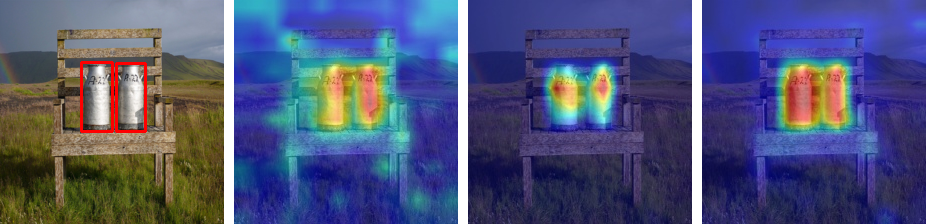}
  \end{tabular}
  \end{center}
    \caption{Visualization examples of the heatmaps on the images from ImageNet ILSVRC 2012.}
    \label{fig:ILSVRC_vis}
\end{figure*}

\begin{figure*}[!t]
  \begin{center}
  \small
  \begin{tabular}{C{0.65in}p{0.65in}C{0.65in}C{0.65in}L{0.03in}p{0.65in}p{0.65in}C{0.65in}C{0.65in}}
        \makecell{bounding\\box} & \makecell{Attention\\Rollout} &  LRP-based & Ours & {} & \makecell{bounding\\box} & \makecell{Attention\\Rollout} & LRP-based & Ours
  \end{tabular}
  \begin{tabular}{C{3.3in}C{3.3in}}
      \includegraphics[width=0.98\linewidth]{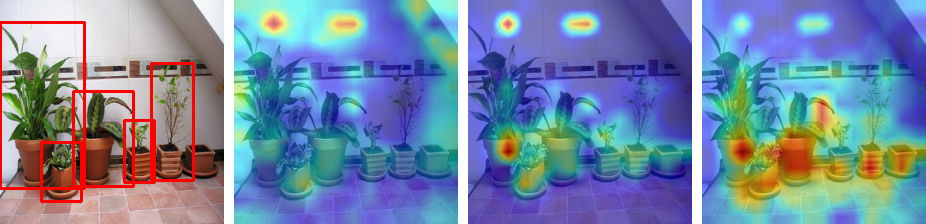} &
      \includegraphics[width=0.98\linewidth]{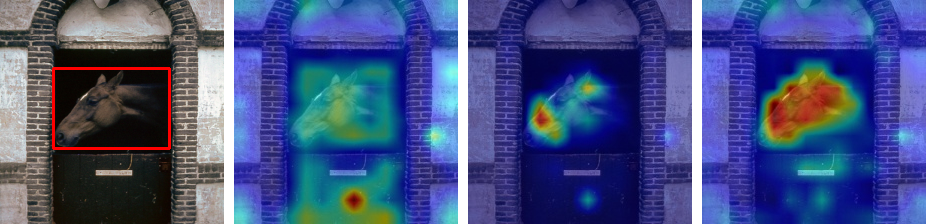}\\
      \includegraphics[width=0.98\linewidth]{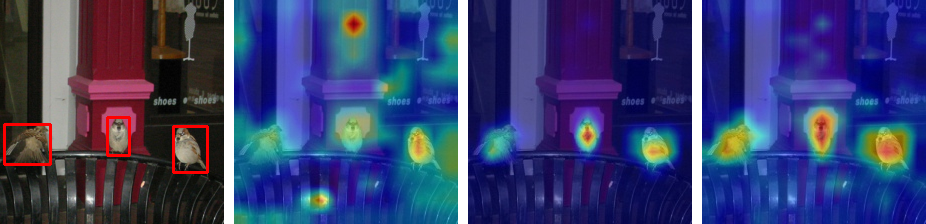} &
      \includegraphics[width=0.98\linewidth]{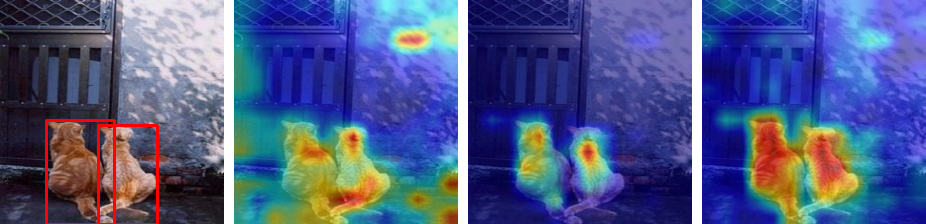}\\
      \includegraphics[width=0.98\linewidth]{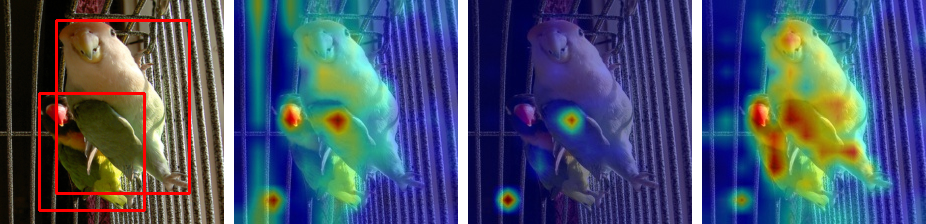} &
      \includegraphics[width=0.98\linewidth]{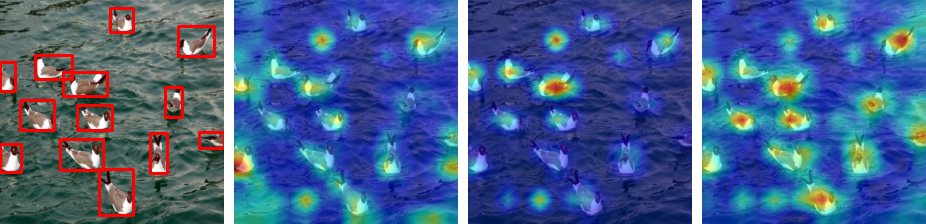}\\
      \includegraphics[width=0.98\linewidth]{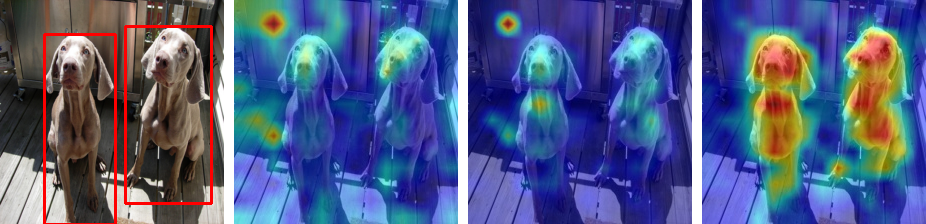} &
      \includegraphics[width=0.98\linewidth]{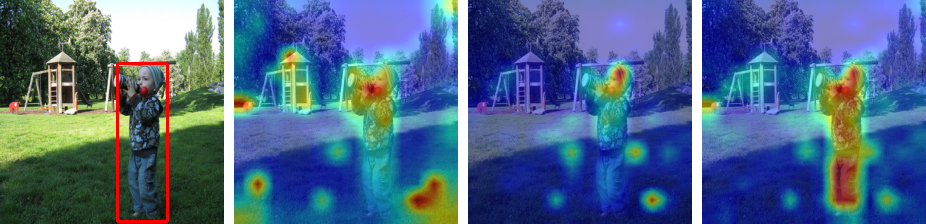}\\
      \includegraphics[width=0.98\linewidth]{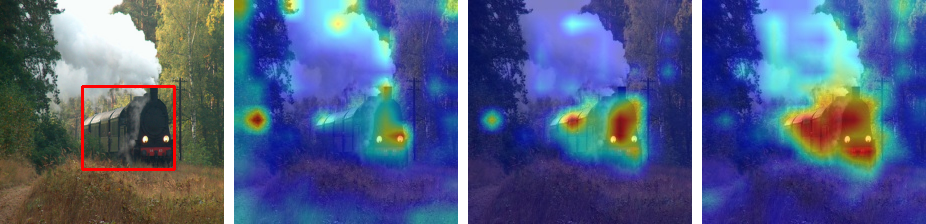} &
      \includegraphics[width=0.98\linewidth]{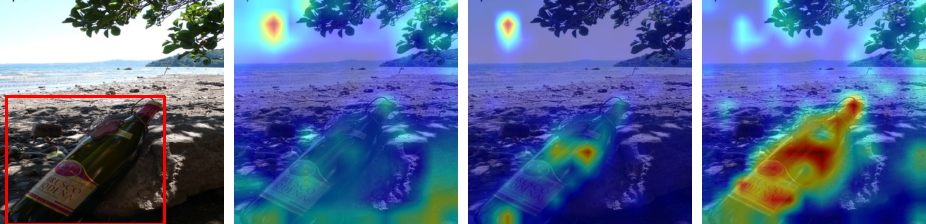}\\
      \includegraphics[width=0.98\linewidth]{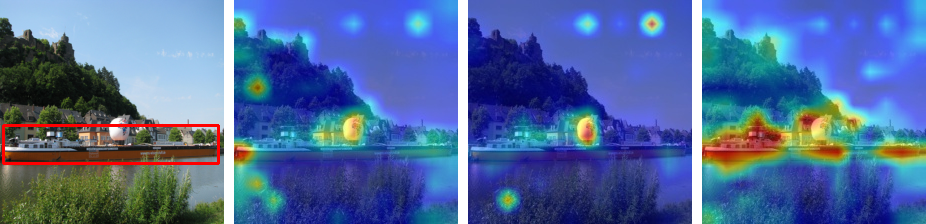} &
      \includegraphics[width=0.98\linewidth]{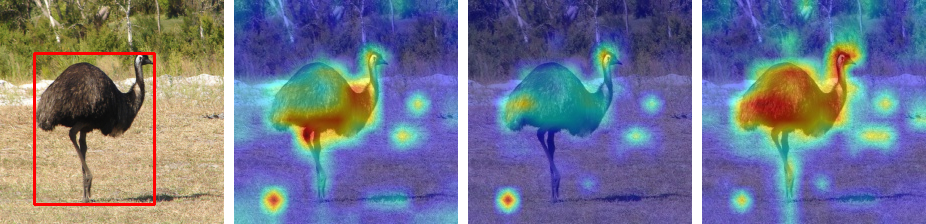}\\
      \includegraphics[width=0.98\linewidth]{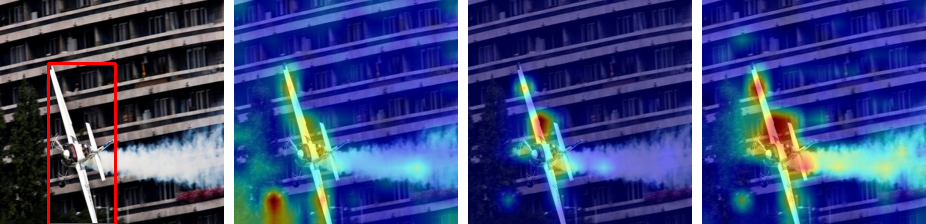} &
      \includegraphics[width=0.98\linewidth]{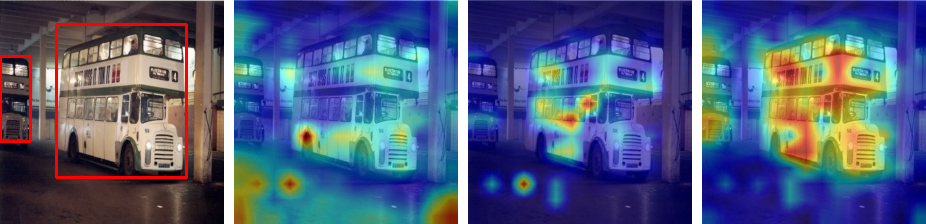}\\
      \includegraphics[width=0.98\linewidth]{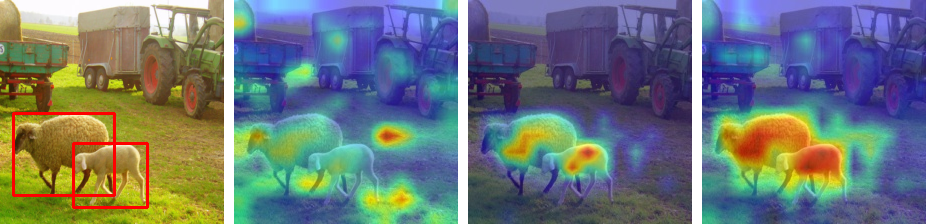} &
      \includegraphics[width=0.98\linewidth]{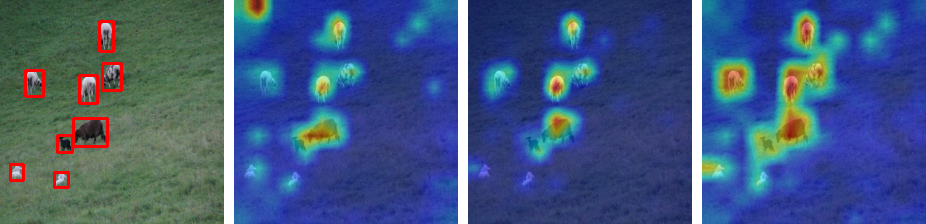}\\
      \includegraphics[width=0.98\linewidth]{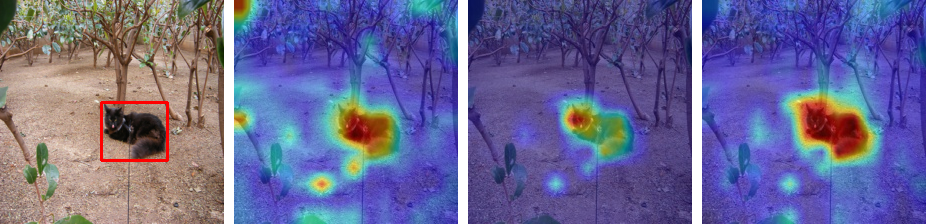} &
      \includegraphics[width=0.98\linewidth]{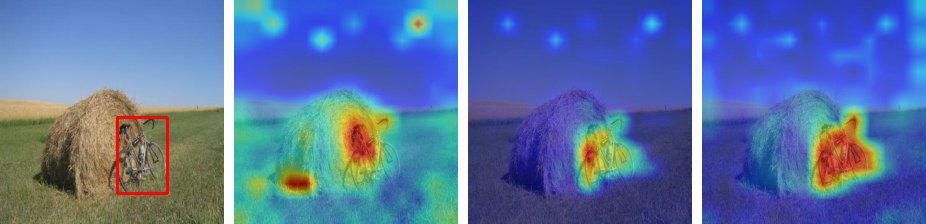}\\
      \includegraphics[width=0.98\linewidth]{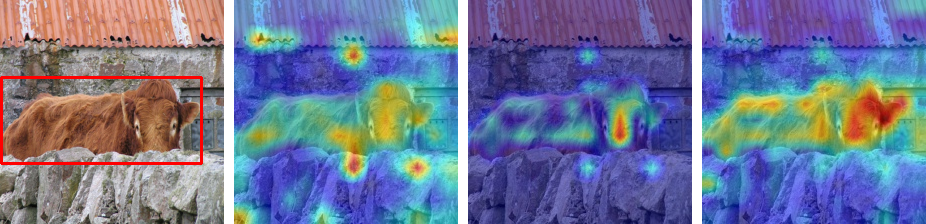} &
      \includegraphics[width=0.98\linewidth]{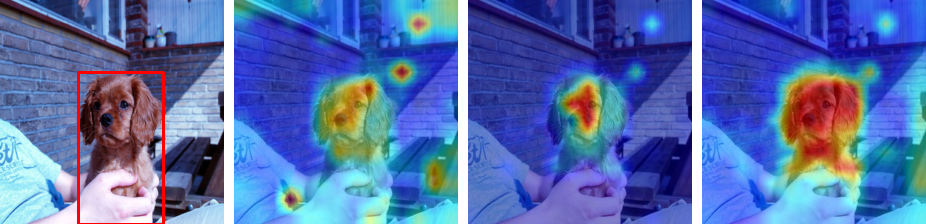}
  \end{tabular}
  \end{center}
    \caption{Visualization examples of the heatmaps on the images from Pascal VOC 2012.}
    \label{fig:Pascal_vis}
\end{figure*}

\begin{figure*}[!t]
  \begin{center}
  \small
  \begin{tabular}{C{0.65in}p{0.65in}C{0.65in}C{0.65in}L{0.03in}p{0.65in}p{0.65in}C{0.65in}C{0.65in}}
        \makecell{bounding\\box} & \makecell{Attention\\Rollout} &  LRP-based & Ours & {} & \makecell{bounding\\box} & \makecell{Attention\\Rollout} & LRP-based & Ours
  \end{tabular}
  \begin{tabular}{C{3.3in}C{3.3in}}
      \includegraphics[width=0.98\linewidth]{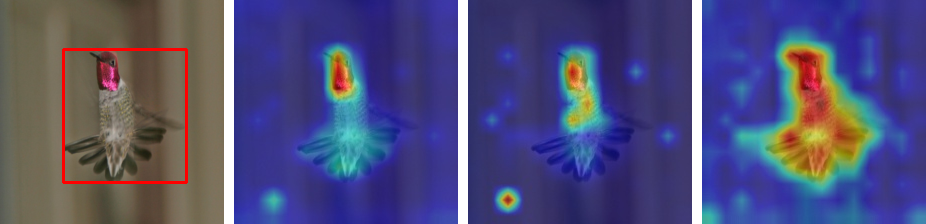} &
      \includegraphics[width=0.98\linewidth]{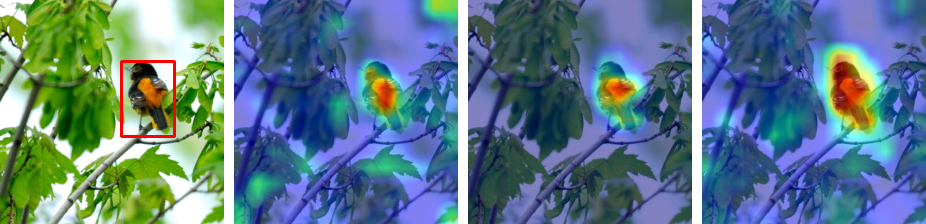}\\
      \includegraphics[width=0.98\linewidth]{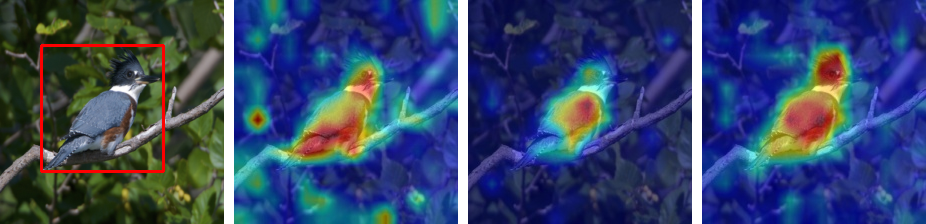} &
      \includegraphics[width=0.98\linewidth]{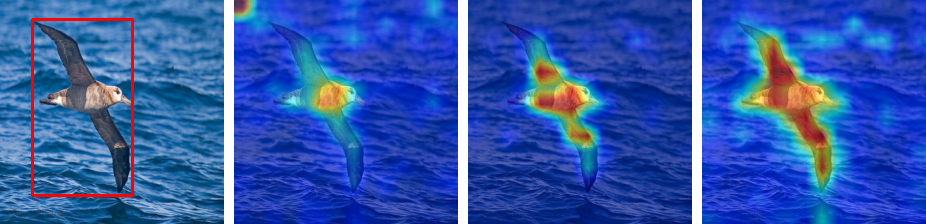}\\
      \includegraphics[width=0.98\linewidth]{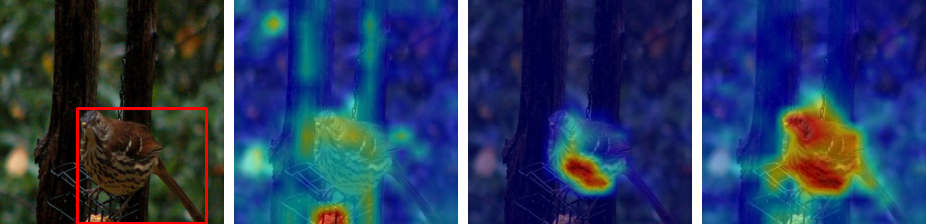} &
      \includegraphics[width=0.98\linewidth]{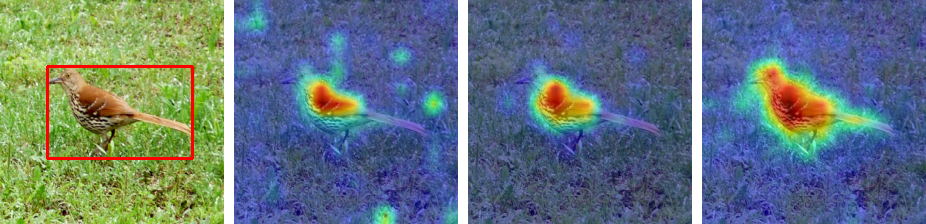}\\
      \includegraphics[width=0.98\linewidth]{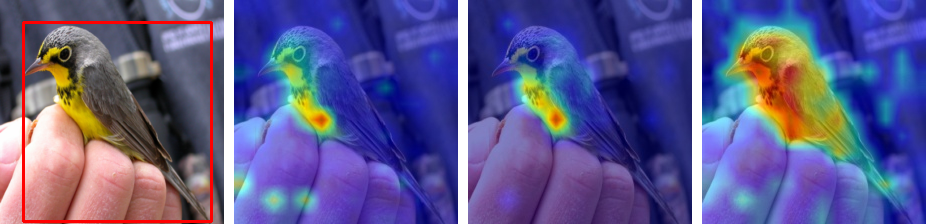} &
      \includegraphics[width=0.98\linewidth]{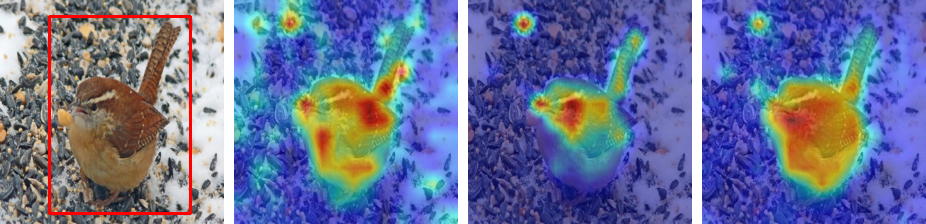}\\
      \includegraphics[width=0.98\linewidth]{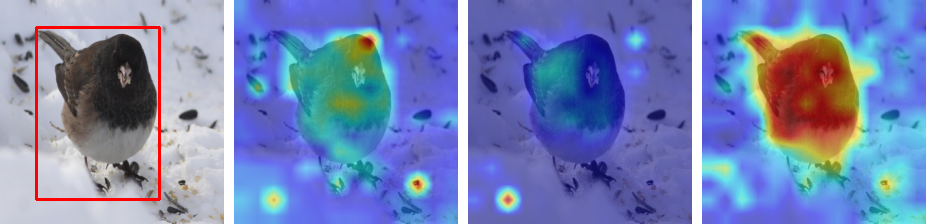} &
      \includegraphics[width=0.98\linewidth]{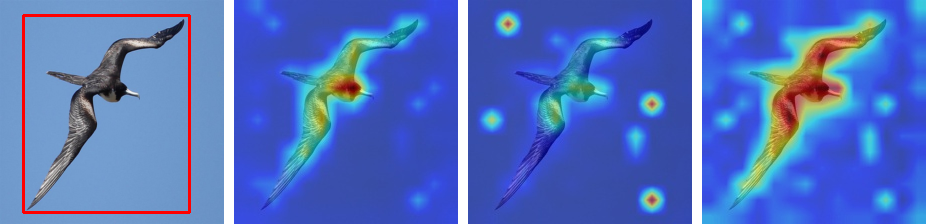}\\
      \includegraphics[width=0.98\linewidth]{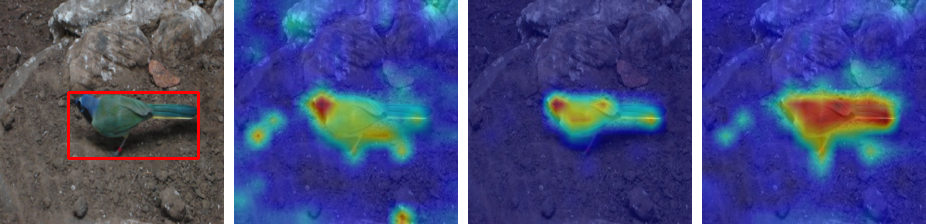} &
      \includegraphics[width=0.98\linewidth]{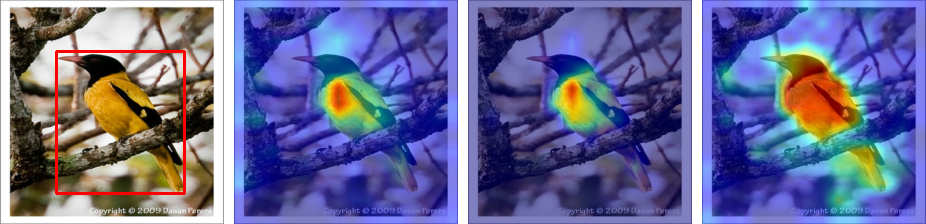}\\
      \includegraphics[width=0.98\linewidth]{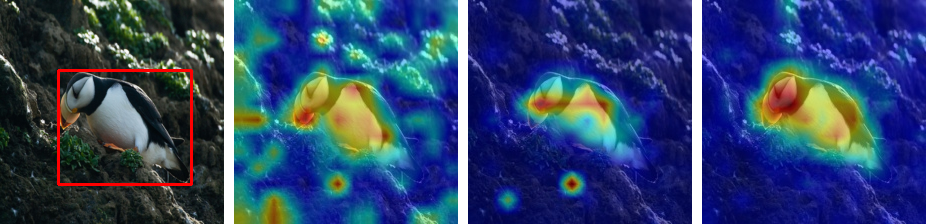} &
      \includegraphics[width=0.98\linewidth]{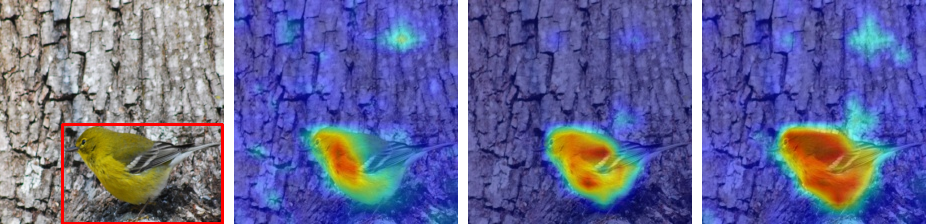}\\
      \includegraphics[width=0.98\linewidth]{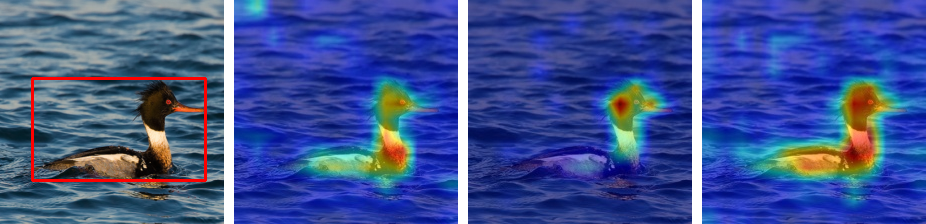} &
      \includegraphics[width=0.98\linewidth]{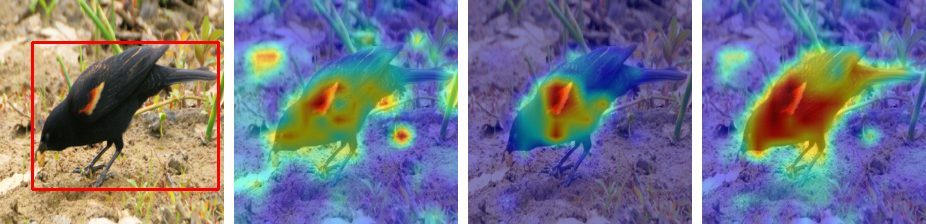}\\
      \includegraphics[width=0.98\linewidth]{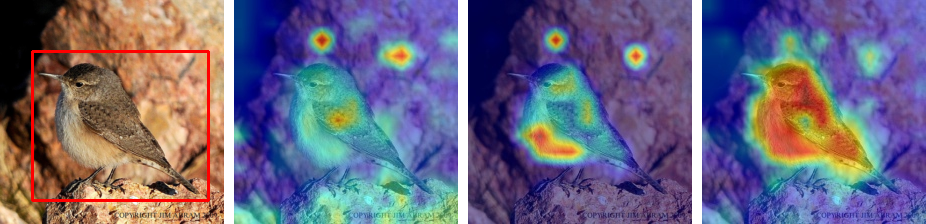} &
      \includegraphics[width=0.98\linewidth]{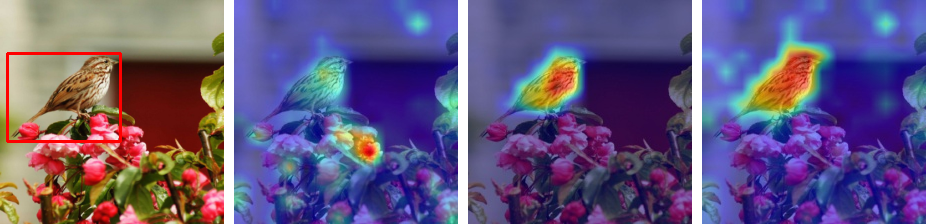}\\
      \includegraphics[width=0.98\linewidth]{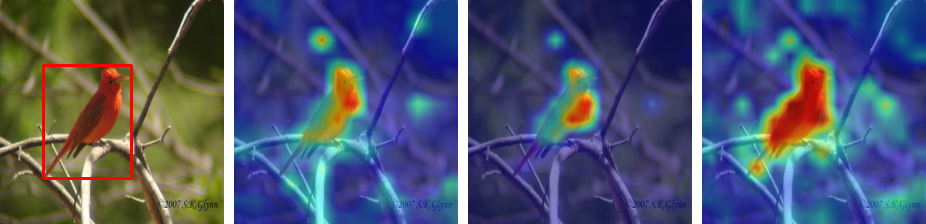} &
      \includegraphics[width=0.98\linewidth]{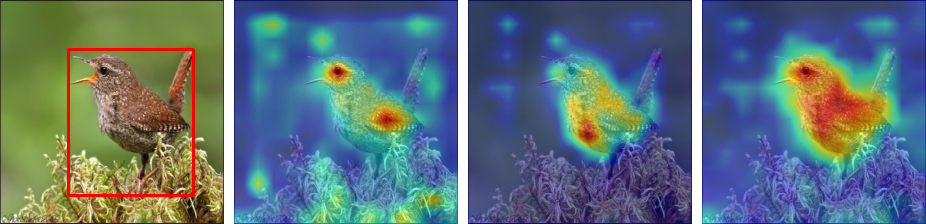}
  \end{tabular}
  \end{center}
    \caption{Visualization examples of the heatmaps on the images from CUB200.}
    \label{fig:CUB200_vis}
\end{figure*}
\end{document}